\documentclass[12pt,twoside]{mitthesis}
\usepackage{lgrind}
\usepackage{cmap}
\usepackage[T1]{fontenc}
\usepackage[utf8]{inputenc}
\usepackage{textcomp}
\usepackage{amssymb,amsmath}
\usepackage{inconsolata}

    \usepackage{color}
    \usepackage{fancyvrb}

    \DefineVerbatimEnvironment{Highlighting}{Verbatim}{commandchars=\\\{\}}
    \newenvironment{Shaded}{}{}
    \newcommand{\KeywordTok}[1]{\textcolor[rgb]{0.00,0.44,0.13}{\textbf{{#1}}}}
    
    \newcommand{\DecValTok}[1]{\textcolor[rgb]{0.25,0.63,0.44}{{#1}}}

    \newcommand{\CharTok}[1]{\textcolor[rgb]{0.25,0.44,0.63}{{#1}}}
    
    \newcommand{\StringTok}[1]{\textcolor[rgb]{0.25,0.44,0.63}{{#1}}}

    \newcommand{\ControlFlowTok}[1]{\textcolor[rgb]{0.00,0.44,0.13}{\textbf{{#1}}}}
    \newcommand{\OperatorTok}[1]{\textcolor[rgb]{0.40,0.40,0.40}{{#1}}}
    \newcommand{\BuiltInTok}[1]{{#1}}

    \newcommand{\NormalTok}[1]{{#1}}
    \usepackage{longtable,booktabs}
    \usepackage{graphicx,grffile}
    \makeatletter
    \def\maxwidth{\ifdim\Gin@nat@width>\linewidth\linewidth\else\Gin@nat@width\fi}
    \def\maxheight{\ifdim\Gin@nat@height>\textheight\textheight\else\Gin@nat@height\fi}
    \makeatother
    \setkeys{Gin}{width=\maxwidth,height=\maxheight,keepaspectratio}

\usepackage[unicode=true]{hyperref}
\hypersetup{breaklinks=true,
            bookmarks=true,
            pdfauthor={William Whitney},
            pdftitle={Disentangled Representations in Neural Models},
            colorlinks=true,
            citecolor=blue,
            urlcolor=blue,
            linkcolor=black,
            pdfborder={0 0 0}}
\providecommand{\tightlist}{%
  \setlength{\itemsep}{0pt}\setlength{\parskip}{0pt}}

\usepackage{upquote}

\usepackage{titlesec}
\titleformat{\chapter}{\normalfont\huge\bfseries}{\thechapter.}{20pt}{\huge\bf}

\begin{document}

\title{Disentangled Representations in Neural Models}

\author{William Whitney}
\prevdegrees{S.B., Massachusetts Institute of Technology (2013)}
\department{Department of Electrical Engineering and Computer Science}

\degree{Master of Engineering in Computer Science and Engineering}

\degreemonth{February}
\degreeyear{2016}
\thesisdate{January 29, 2016}


\supervisor{Joshua B. Tenenbaum}{Professor}

\chairman{Christopher Terman}{Chairman, Masters of Engineering Thesis Committee}

\maketitle



\cleardoublepage
\setcounter{savepage}{\thepage}
\begin{abstractpage}
Representation learning is the foundation for the recent success of
neural network models. However, the distributed representations
generated by neural networks are far from ideal. Due to their highly
entangled nature, they are difficult to reuse and interpret, and they do
a poor job of capturing the sparsity which is present in real-world
transformations.

In this paper, I describe methods for learning \emph{disentangled}
representations in the two domains of graphics and computation. These
methods allow neural methods to learn representations which are easy to
interpret and reuse, yet they incur little or no penalty to performance.
In the Graphics section, I demonstrate the ability of these methods to
infer the generating parameters of images and rerender those images
under novel conditions. In the Computation section, I describe a model
which is able to factorize a multitask learning problem into subtasks
and which experiences no catastrophic forgetting. Together these
techniques provide the tools to design a wide range of models that learn
disentangled representations and better model the factors of variation
in the real world.
\end{abstractpage}


\cleardoublepage

\section*{Acknowledgments}
I would like to thank my girlfriend, Benjana, who is my joy. She lets me
see my work through new eyes.

I thank Tejas Kulkarni for his mentorship and friendship. He has been my
gateway into this world and a strong guiding influence, and I would not
be here without him.

I thank Josh Tenenbaum for his guidance. He has pushed me to think about
the most fundamental problems.

I thank my parents for their constant love and support. They made me
this way, so please direct all complaints to them.

I thank Thomas Vetter for access to the Basel face model. I am grateful
for support from the MIT Center for Brains, Minds, and Machines (CBMM).

%


    {
    \setcounter{tocdepth}{2}
    \tableofcontents
    }
    \listoffigures

\chapter{Introduction}\label{introduction}

Representation is one of the most fundamental problems in machine
learning. It underlies such varied fields as vision, speech recognition,
natural language processing, reinforcement learning, and graphics. Yet
the question of what makes a good representation is a deceptively
complex one. On the one hand, we would like representations which
perform well on real-world tasks. On the other, we would like to be able
to interpret these representations, and they should be useful for tasks
beyond those explicit in their initial design.

Presently representations come in two varieties: those that are
designed, and those that are learned from data. Designed representations
can perfectly match our desire for structured reuse and
interpretability, while learned representations require no expert
knowledge yet outperform designed features on practically every task
that has sufficient data.

This tension has been the source of great debate in the community.
Clearly a representation which has some factorizable structure can be
more readily reused in part or in whole for some new task. Much more
than being just an issue of interpretation, this concern has a very
practical focus on generalization; it is unreasonable to spend a
tremendous amount of data building a new representation for every single
task, even when those tasks have strong commonalities. Since we have
knowledge about the true structure of the problem, we can design a
representation which is factorized and thus reusable.

Learned representations take a very different approach to the problem.
Instead of attempting to incorporate expert knowledge of a domain to
create a representation which will be broadly useful, a learned
representation is simply the solution to a single optimization problem.
It is custom-designed to solve precisely the task it was trained on, and
while it may be possible to reverse-engineer such a representation for
reuse elsewhere, it is typically unclear how to do so and how useful it
will be in a secondary setting.

Despite the obvious advantages of structured representations, those we
design are inherently limited by our understanding of the problem.
Perhaps it is possible to design image features with spokes and circles
that will be able to distinguish a bike wheel from a car wheel, but
there are a million subtle clues that no human would think to include.
As a result, in domain after domain, as the datasets have grown larger,
representations learned by deep neural networks have come to dominate.

The dominance of optimization-based representation learning is
unavoidable and in fact hugely beneficial to the field. However, the
weaknesses of these learned representations is not inherent in their
nature; it merely reflects the limits of our current tasks and
techniques.

This thesis represents an effort to bring together the advantages of
each of these techniques to learn representations which perform well,
yet have valuable structure. Using the two domains of graphics and
programs, I will discuss the rationale, techniques, and results of
bringing structure to neural representations.

\section{Document overview}\label{document-overview}

The next chapter discusses various criteria for assessing the quality of
a representation.

In the following two chapters, I use these criteria to discuss
representations in the domains of graphics and computer programs. Each
chapter begins with an overview of the problems in the field and related
work, then moves on to a description of the specific problem I address,
my methods, and the results.

In the final chapter I discuss the significance of this work to the
field and promising directions for further research.

\chapter{Desiderata for
representations}\label{desiderata-for-representations}

When evaluating a representation, it is valuable to have a clear set of
goals. Several of the goals stated here have substantial overlap, and to
some degree a representation which perfectly addresses one may
automatically fulfill another as well. However, each of them provides a
distinct benefit, and their significance must be considered with respect
to those benefits.

\section{Disentangled}\label{disentangled}

A representation which is \emph{disentangled} for a particular dataset
is one which is sparse over the transformations present in that data
(Bengio, Courville, and Vincent
\protect\hyperlink{ref-bengio2013representation}{2013}). For example,
given a dataset of indoor videos, a representation that explicitly
represents whether or not the lights are on is more disentangled than a
representation composed of raw pixels. This is because for the common
transformation of flipping the light switch, the first representation
will only change in only that single dimension (light on or off),
whereas the second will change in every single dimension (pixel).

For a representation to be disentangled implies that it factorizes some
latent cause or causes of variation. If there are two causes for the
transformations in the data which do not always happen together and
which are distinguishable, a maximally disentangled representation will
have a structure that separates those causes. In the indoor scenes
example above, there might be two sources of lighting: sunlight and
electric lights. Since transformations in each of these latent variables
occur independently, it is more sparse to represent them separately.

The most disentangled possible representation for some data is a
graphical model expressing the ``true'' generative process for that
data. In graphics this model might represent each object in a room, with
its pose in the scene and its intrinsic reflectance characteristics, and
the sources of lighting. For real-world transformations involving
motion, only the pose of each object needs to be updated. As the
lighting shifts, nothing about the representation of the objects needs
to be changed; the visual appearance of the object can be recalculated
from the new lighting variables.

In a certain light, all of science is one big unsupervised learning
problem in which we search for the most disentangled representation of
the world around us.

\section{Interpretable}\label{interpretable}

An \emph{interpretable} representation is, simply enough, one that is
easy for humans to understand. A human should be able to make
predictions about what changes in the source domain would do in the
representation domain and vice versa. In a graphics engine's
representation of a scene, for example, it is easy for a person to
predict things like ``What would the image (source domain) look like if
I changed the angle of this chair (representation domain) by 90°?'' By
contrast, in the representation of a classically-trained autoencoder, it
is practically impossible for a person to visualize the image that would
be generated if some particular component were changed.

Interpretability is closely related with disentanglement. This is
because, in ``human'' domains of data like vision and audition, humans
are remarkably good at inferring generative structure, and tend to
internally use highly disentangled representations. However, this
relationship only holds for datasets which are similar to human
experience. One could construct a dataset of videos in which the most
common transformation between frames was for each pixel in the image to
change its hue by an amount proportional to the number of characters in
the Arabic name of the object shown in that pixel. The most disentangled
representation of these videos would perfectly match this structure, but
this disentangled representation would be less interpretable than a
table of English names of objects and how much their color changes per
frame.

In a real-world setting, the most disentangled possible representation
of stock market prices might involve a latent which represents a complex
agglomeration of public opinion from the news, consumer confidence
ratings, and estimates of the Fed's likelihood of raising rates. Such a
latent might truly be the best and most independent axis of variation
for predicting the stock price, yet it would not be as easy to interpret
as a representation with one latent for public opinion, one latent for
the Fed, and one latent for consumer confidence. In such a non-human
domain, our intuitions about the factors of variation may not hold, and
as a result the representations that make sense to us and those that
accurately represent the factors of variation may diverge.

Interpretability is extremely valuable in many domains. If a doctor is
attempting to plan a course of treatment for a patient, they need to be
able to reason about the factors in a diagnostic model they're using.
Even if an algorithm doesn't need to interface with a human at runtime,
it's very hard to debug a system during development if you don't
understand what it's doing.

\section{Performant}\label{performant}

A \emph{performant} representation for a task contains the information
needed to perform well on that task.

If the task is determining whether or not there is a dog in a room, a
representation consisting of a photograph of the room would be less
performant than a 3D voxel model of the room, which in turn would be
less performant than a single binary bit representing whether or not
there is a dog.

\section{Reusable}\label{reusable}

A \emph{reusable} representation is one that is performant for many
tasks in the same domain.

To continue the example above, a 3D voxel representation of a room is
more reusable than one indicating whether or not the room contains a
dog. Somewhere in between the two would be a representation consisting
of the facts,

\begin{itemize}
\tightlist
\item
  Is there an animal?
\item
  Is it furry?
\item
  How big is it?
\item
  What color is it?
\end{itemize}

This representation would be able to solve the task of whether or not
the room contains a dog with high probability, and would also be able to
inform the task of whether or not the room contains a gorilla, or a
whale. However, the tasks it can address are a strict subset of the
voxel representation, which could also solve such problems as ``Where is
the couch?''

\section{Compact}\label{compact}

A \emph{compact} representation is one which occupies few bits.

Compactness may not seem inherently important; it is typically
irrelevant if the representation of an image takes up one megabyte or
two. However, compactness provides a very valuable forcing function. One
might build a weather forecasting model which represents the state of
the world down to the very last butterfly.

The actions of this butterfly might be indeterminate given the other
latents in the model, so it is disentangled; it might be perfectly easy
to understand the meaning of the butterfly's representation, so it is
interpretable; it might be valuable in some other system or context, so
it is reusable; and it might even minutely improve the performance of
the weather forecast, so it is performant. But somehow none of this
quite justifies its presence in a weather model.

Compactness says that we only care about the most important factors of
variation.

\chapter{Disentanglement in Vision}\label{disentanglement-in-vision}

\section{Introduction}\label{introduction-1}

Deep learning has led to remarkable breakthroughs in learning
hierarchical representations from images. Models such as Convolutional
Neural Networks (CNNs) (LeCun and Bengio
\protect\hyperlink{ref-lecun1995convolutional}{1995}), Restricted
Boltzmann Machines, (Hinton, Osindero, and Teh
\protect\hyperlink{ref-hinton2006fast}{2006}, Salakhutdinov and Hinton
(\protect\hyperlink{ref-salakhutdinov2009deep}{2009})), and
Auto-encoders (Bengio \protect\hyperlink{ref-bengio2009learning}{2009},
Vincent et al. (\protect\hyperlink{ref-vincent2010stacked}{2010})) have
been successfully applied to produce multiple layers of increasingly
abstract visual representations. However, there is relatively little
work on characterizing the optimal representation of the data. While
Cohen et al. (\protect\hyperlink{ref-cohen2014learning}{2014}) have
considered this problem by proposing a theoretical framework to learn
irreducible representations with both invariances and equivariances,
coming up with the best representation for any given task is an open
question.

To shed some light on this question, let us consider our list of desires
for representations in the specific context of vision.

\begin{enumerate}
\def\labelenumi{\arabic{enumi}.}
\item
  \textbf{Disentangled}: When applied to real-world transformations over
  images, i.e.~video, a good representation will change only sparsely.
  That is, the expectation over a set of videos of the number of
  dimensions of the representation which change between each frame
  should be small. In practice this means that common transformations,
  like movement of an object, should be expressed concisely; whereas
  uncommon transformations, like a solid object turning inside out, may
  be more expensive to express.
\item
  \textbf{Interpretable}: To be highly interpretable, a representation
  needs to line up well with the one that's in our heads. It should
  express objects separately from their conditions, and common
  transformations should be monotonic and smooth in representation
  space.
\item
  \textbf{Performant}: Representations of visual content need to be
  quite rich; in order to be able to solve problems like ``Which of
  these objects is in front of the other?'' a representation must
  understand much more than just pixels.
\item
  \textbf{Reusable}: A representation which was learned to solve a very
  specific task, like perhaps that of the DQN (Mnih et al.
  \protect\hyperlink{ref-mnih2015human}{2015}), will not be helpful in
  other settings, like the real world. To be reusable a model needs to
  capture the structure that is universal to our world.
\item
  \textbf{Compact}: Representations of images should be able to
  efficiently compress the images in their domain.
\end{enumerate}

The ``vision as inverse graphics'' paradigm suggests a representation
for images which provides these features. Computer graphics consists of
a function to go from compact descriptions of scenes (the \emph{graphics
code}) to images, and this graphics code is typically disentangled to
allow for rendering scenes with fine-grained control over
transformations such as object location, pose, lighting, texture, and
shape. This encoding is designed to easily and interpretably represent
sequences of real data so that common transformations may be compactly
represented in software code; this criterion is conceptually identical
to disentanglement, and graphics codes conveniently align with the
properties of an ideal representation. Graphics codes are the
representations which we as a society have designed to be the single
most general-purpose, interpretable, and generally usable way to express
scenes.

Early work by Tenenbaum et al.
(\protect\hyperlink{ref-tenenbaum2000separating}{2000}) was among the
first to explore this idea, and used bilinear models to differentiate
between extrinsic and intrinsic factors in images. Recent work in
inverse graphics (Mansinghka et al.
\protect\hyperlink{ref-mansinghka2013approximate}{2013}, Kulkarni et al.
(\protect\hyperlink{ref-kulkarni2014inverse}{2014}), Kulkarni, Kohli, et
al. (\protect\hyperlink{ref-kulkarni2015picture}{2015})) follows a
general strategy of defining a probabilistic with latent parameters,
then using an inference algorithm to find the most appropriate set of
latent parameters given the observations. Tieleman et al.
(\protect\hyperlink{ref-tieleman2014optimizing}{2014}) moved beyond this
two-stage pipeline by using a generic encoder network and a
domain-specific decoder network to approximate a 2D rendering function.
However, none of these approaches have been shown to automatically
produce a semantically-interpretable graphics code and to learn a 3D
rendering engine to reproduce images.

I present an approach, first described in (Kulkarni, Whitney, et al.
\protect\hyperlink{ref-kulkarni2015deep}{2015}), which attempts to learn
interpretable graphics codes for complex transformations such as
out-of-plane rotations and lighting variations. Given a set of images,
we use a hybrid encoder-decoder model to learn a representation that is
disentangled with respect to various transformations such as object
out-of-plane rotations and lighting variations. We employ a deep
directed graphical model with many layers of convolution and
de-convolution operators that is trained using the Stochastic Gradient
Variational Bayes (SGVB) algorithm (Kingma and Welling
\protect\hyperlink{ref-kingma2013auto}{2013}).

We propose a training procedure to encourage each group of neurons in
the graphics code layer to distinctly represent a specific
transformation. To learn a disentangled representation, we train using
data where each mini-batch has a set of active and inactive
transformations, but we do not provide target values as in supervised
learning; the objective function remains reconstruction quality. For
example, a nodding face would have the 3D elevation transformation
active but its shape, texture and other transformations would be
inactive. We exploit this type of training data to force chosen neurons
in the graphics code layer to specifically represent active
transformations, thereby automatically creating a disentangled
representation. Given a single face image, our model can re-generate the
input image with a different pose and lighting. We present qualitative
and quantitative results of the model's efficacy at learning a 3D
rendering engine.

\section{Related Work}\label{related-work}

As mentioned previously, a number of generative models have been
proposed in the literature to obtain abstract visual representations.
Unlike most RBM-based models (Hinton, Osindero, and Teh
\protect\hyperlink{ref-hinton2006fast}{2006}, Salakhutdinov and Hinton
(\protect\hyperlink{ref-salakhutdinov2009deep}{2009}), Lee et al.
(\protect\hyperlink{ref-lee2009convolutional}{2009})), our approach is
trained using back-propagation with objective function consisting of
data reconstruction and the variational bound.

Relatively recently, Kingma et al. (Kingma and Welling
\protect\hyperlink{ref-kingma2013auto}{2013}) proposed the SGVB
algorithm to learn generative models with continuous latent variables.
In this work, a feed-forward neural network (encoder) is used to
approximate the posterior distribution and a decoder network serves to
enable stochastic reconstruction of observations. In order to handle
fine-grained geometry of faces, we work with relatively large scale
images ($150 \times 150$ pixels). Our approach extends and applies the
SGVB algorithm to jointly train and utilize many layers of convolution
and de-convolution operators for the encoder and decoder network
respectively. The decoder network is a function that transform a compact
graphics code (200 dimensions) to a $150 \times 150$ image. We propose
using unpooling (nearest neighbor sampling) followed by convolution to
handle the massive increase in dimensionality with a manageable number
of parameters.

(Dosovitskiy, Springenberg, and Brox
\protect\hyperlink{ref-dosovitskiy2015learning}{2015}) proposed using
CNNs to generate images given object-specific parameters in a supervised
setting. As their approach requires ground-truth labels for the graphics
code layer, it cannot be directly applied to image interpretation tasks.
Our work is similar to Ranzato et al.
(\protect\hyperlink{ref-ranzato2007unsupervised}{2007}), whose work was
amongst the first to use a generic encoder-decoder architecture for
feature learning. However, in comparison to our proposal their model was
trained layer-wise, the intermediate representations were not
disentangled like a graphics code, and their approach does not use the
variational auto-encoder loss to approximate the posterior distribution.
Our work is also similar in spirit to (Tang, Salakhutdinov, and Hinton
\protect\hyperlink{ref-tang2012deep}{2012}), but in comparison our model
does not assume a Lambertian reflectance model and implicitly constructs
the 3D representations. Another piece of related work is Desjardins et
al. (\protect\hyperlink{ref-desjardins2012disentangling}{2012}), who
used a spike and slab prior to factorize representations in a generative
deep network.

Quite recently, (Jaderberg et al.
\protect\hyperlink{ref-jaderberg2015spatial}{2015}) proposes a model
which explicitly captures the pose of objects in a scene through the use
of predefined 2D affine transformations, which leads pose and identity
to be disentangled. (Mansimov et al.
\protect\hyperlink{ref-mansimov2015generating}{2015}) use an attention
mechanism to generate images from text; this approach has the potential
to learn functions which are parametrized by highly disentangled
symbolic representations. (Theis and Bethge
\protect\hyperlink{ref-theis2015generative}{2015}) use spatial LSTMs to
build generative models of textures in natural images.

In comparison to prior approaches, it is important to note that our
encoder network produces the interpretable and disentangled
representations necessary to learn a meaningful 3D graphics engine. A
number of inverse-graphics inspired methods have recently been proposed
in the literature (Mansinghka et al.
\protect\hyperlink{ref-mansinghka2013approximate}{2013}). However, most
such methods rely on hand-crafted rendering engines. The exception to
this is work by Hinton et al.
(\protect\hyperlink{ref-hinton2011transforming}{2011}) and Tieleman
(\protect\hyperlink{ref-tieleman2014optimizing}{2014}) on
\emph{transforming autoencoders} which use a domain-specific decoder to
reconstruct input images.

(Yang et al. \protect\hyperlink{ref-yang2015weakly}{2015}) follows up on
our work with a recurrent model which learns similar disentangled
representations from watching synthesized video.

\section{Model}\label{model}

\begin{figure}[htbp]
\centering
\includegraphics{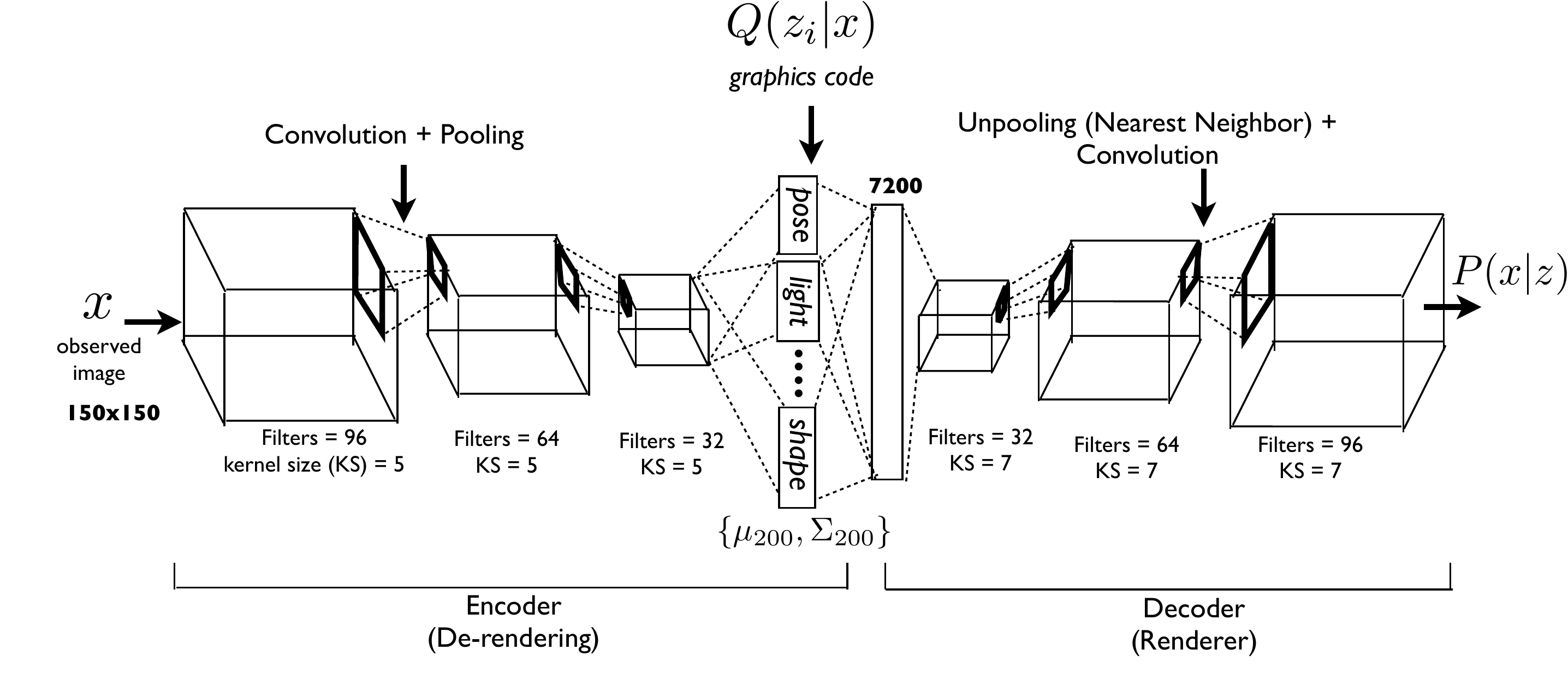}
\caption{\label{fig:overview}\textbf{Model Architecture.} Deep
Convolutional Inverse Graphics Network (DC-IGN) has an encoder and a
decoder. We follow the variational autoencoder (Kingma and Welling
\protect\hyperlink{ref-kingma2013auto}{2013}) architecture with
variations. The encoder consists of several layers of convolutions
followed by max-pooling and the decoder has several layers of unpooling
(upsampling using nearest neighbors) followed by convolution. (a) During
training, data $x$ is passed through the encoder to produce the
posterior approximation $Q(z_i|x)$, where $z_i$ consists of scene
latent variables such as pose, light, texture or shape. In order to
learn parameters in DC-IGN, gradients are back-propagated using
stochastic gradient descent using the following variational object
function: $-log(P(x|z_i)) + KL(Q(z_i|x)||P(z_i))$ for every $z_i$.
We can force DC-IGN to learn a disentangled representation by showing
mini-batches with a set of inactive and active transformations
(e.g.~face rotating, light sweeping in some direction etc). (b) During
test, data $x$ can be passed through the encoder to get latents
$z_i$. Images can be re-rendered to different viewpoints, lighting
conditions, shape variations, etc by setting the appropriate graphics
code group $z_i$, which is how one would manipulate an off-the-shelf
3D graphics engine.}
\end{figure}

As shown in Figure fig.~\ref{fig:overview}, the basic structure of the
Deep Convolutional Inverse Graphics Network (DC-IGN) consists of two
parts: an encoder network which captures a distribution over graphics
codes $Z$ given data $x$ and a decoder network which learns a
conditional distribution to produce an approximation $\hat{x}$ given
$Z$. $Z$ can be a disentangled representation containing a factored
set of latent variables $z_i \in Z$ such as pose, light and shape.
This is important in learning a meaningful approximation of a 3D
graphics engine and helps tease apart the generalization capability of
the model with respect to different types of transformations.

\begin{figure}[htbp]
\centering
\includegraphics[width=0.70000\textwidth]{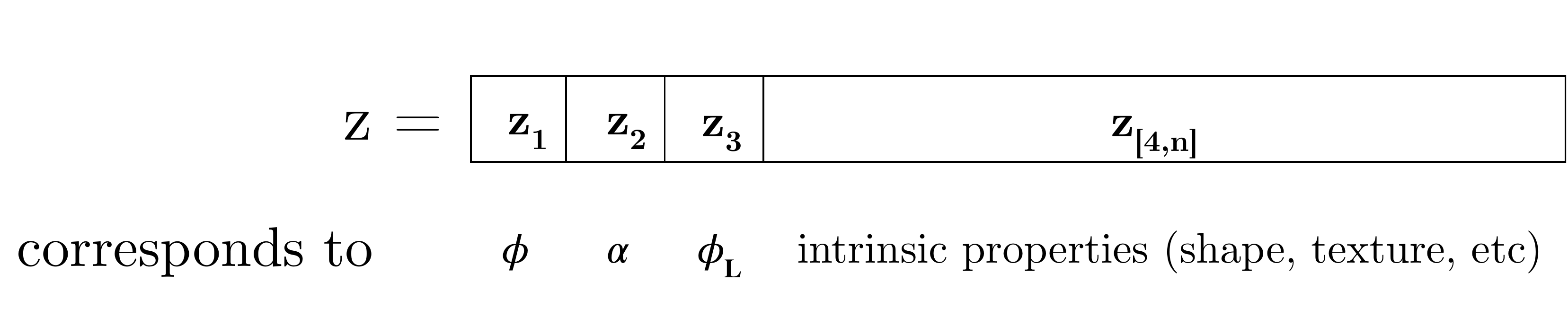}
\caption{\label{fig:latentslegend}\textbf{Structure of the
representation vector.} $\phi$ is the azimuth of the face, $\alpha$
is the elevation of the face with respect to the camera, and $\phi_L$
is the azimuth of the light source.}
\end{figure}

Let us denote the encoder output of DC-IGN to be $y_e = encoder(x)$.
The encoder output is used to parametrize the variational approximation
$Q(z_i|y_e)$, where $Q$ is chosen to be a multivariate normal
distribution. There are two reasons for using this parametrization:

\begin{enumerate}
\def\labelenumi{\arabic{enumi}.}
\tightlist
\item
  Gradients of samples with respect to parameters $\theta$ of $Q$
  can be easily obtained using the reparametrization trick proposed in
  (Kingma and Welling \protect\hyperlink{ref-kingma2013auto}{2013})
\item
  Various statistical shape models trained on 3D scanner data such as
  faces have the same multivariate normal latent distribution (Paysan et
  al. \protect\hyperlink{ref-paysan2009face}{2009}).
\end{enumerate}

Given that model parameters $W_e$ connect $y_e$ and $z_i$, the
distribution parameters $\theta = (\mu_{z_i}, \Sigma_{z_i})$ and
latents $Z$ can then be expressed as:

\[\mu_{z} = W_e  y_e\]

\[\Sigma_{z} = \text{diag}(\exp(W_e  y_e))\]

\[\forall{i}, z_i \sim \mathcal{N}(\mu_{z_i}, \Sigma_{z_i})\]

We present a novel training procedure which allows networks to be
trained to have disentangled and interpretable representations.

\subsection{Training with Specific
Transformations}\label{sec:specifictransforms}

\begin{figure}[htbp]
\centering
\includegraphics{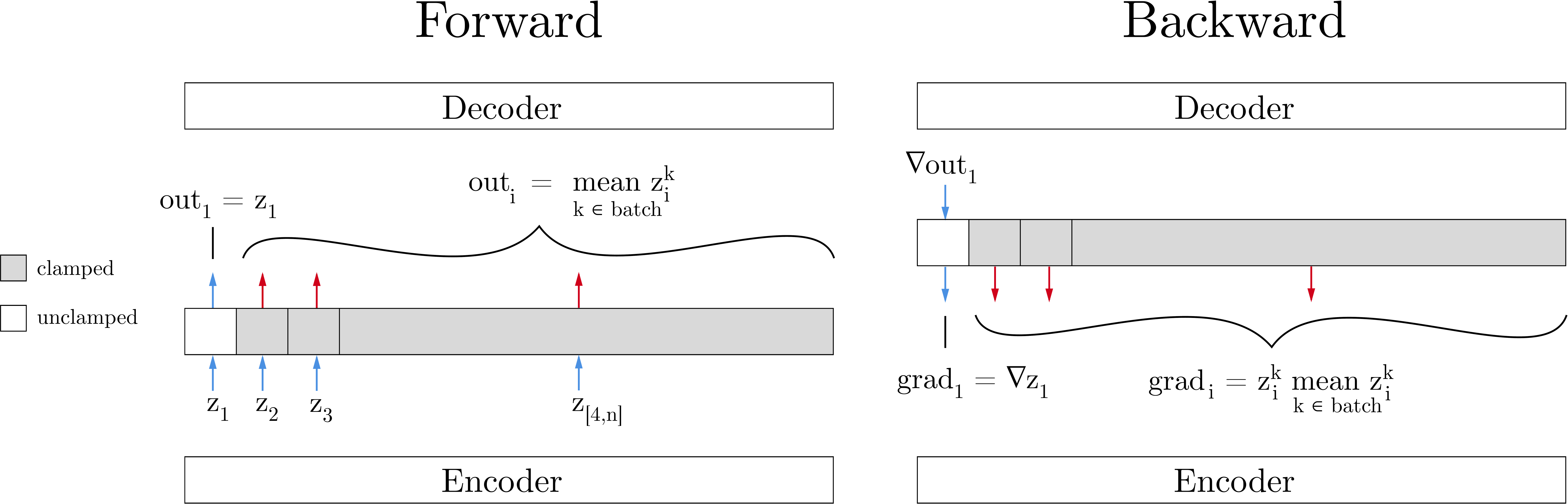}
\caption{\label{fig:selectivetraining}\textbf{Training on a minibatch in
which only $\phi$, the azimuth angle of the face, changes.} During the
forward step, the output from each component $z_i \neq z_1$ of the
encoder is altered to be the same for each sample in the batch. This
reflects the fact that the generating variables of the image (e.g.~the
identity of the face) which correspond to the desired values of these
latents are unchanged throughout the batch. By holding these outputs
constant throughout the batch, the single neuron $z_1$ is forced to
explain all the variance within the batch, i.e.~the full range of
changes to the image caused by changing $\phi$. During the backward
step $z_1$ is the only neuron which receives a gradient signal from
the attempted reconstruction, and all $z_i \neq z_1$ receive a signal
which nudges them to be closer to their respective averages over the
batch. During the complete training process, after this batch, another
batch is selected at random; it likewise contains variations of only one
of ${\phi, \alpha, \phi_L, intrinsic}$; all neurons which do not
correspond to the selected latent are clamped; and the training
proceeds.}
\end{figure}

The main goal of this work is to learn a representation of the data
which consists of disentangled and semantically interpretable latent
variables. We would like only a small subset of the latent variables to
change for sequences of inputs corresponding to real-world events.

One natural choice of target representation for information about scenes
is that already designed for use in graphics engines. If we can
deconstruct a face image by splitting it into variables for pose, light,
and shape, we can trivially represent the same transformations that
these variables are used for in graphics applications.
Fig.~\ref{fig:latentslegend} depicts the representation which we will
attempt to learn.

With this goal in mind, we perform a training procedure which directly
targets this definition of disentanglement. We organize our data into
mini-batches corresponding to changes in only a single scene variable
(azimuth angle, elevation angle, azimuth angle of the light source);
these are transformations which might occur in the real world. We will
term these the \emph{extrinsic} variables, and they are represented by
the components $z_{1,2,3}$ of the encoding.

We also generate mini-batches in which the three extrinsic scene
variables are held fixed but all other properties of the face change.
That is, these batches consist of many different faces under the same
viewing conditions and pose. These \emph{intrinsic} properties of the
model, which describe identity, shape, expression, etc., are represented
by the remainder of the latent variables $z_{[4,200]}$. These
mini-batches varying intrinsic properties are interspersed
stochastically with those varying the extrinsic properties.

We train this representation using SGVB, but we make some key
adjustments to the outputs of the encoder and the gradients which train
it. The procedure (Fig.~\ref{fig:selectivetraining}) is as follows.

\begin{enumerate}
\def\labelenumi{\arabic{enumi}.}
\tightlist
\item
  Select at random a latent variable $z_{train}$ which we wish to
  correspond to one of \{azimuth angle, elevation angle, azimuth of
  light source, intrinsic properties\}.
\item
  Select at random a mini-batch in which that only that variable
  changes.
\item
  Show the network each example in the minibatch and capture its latent
  representation for that example $z^k$.
\item
  Calculate the average of those representation vectors over the entire
  batch.
\item
  Before putting the encoder's output into the decoder, replace the
  values $z_i \neq z_{train}$ with their averages over the entire
  batch. These outputs are ``clamped''.
\item
  Calculate reconstruction error and backpropagate as per SGVB in the
  decoder.
\item
  Replace the gradients for the latents $z_i \neq z_{train}$ (the
  clamped neurons) with their difference from the mean (see
  Sec.~\ref{sec:targetedinvar}). The gradient at $z_{train}$ is passed
  through unchanged.
\item
  Continue backpropagation through the encoder using the modified
  gradient.
\end{enumerate}

Since the intrinsic representation is much higher-dimensional than the
extrinsic ones, it requires more training. Accordingly we select the
type of batch to use in a ratio of about 1:1:1:10, azimuth : elevation :
lighting : intrinsic; we arrived at this ratio after extensive testing,
and it works well for both of our datasets.

This training procedure works to train both the encoder and decoder to
represent certain properties of the data in a specific neuron. By
clamping the output of all but one of the neurons, we force the decoder
to recreate all the variation in that batch using only the changes in
that one neuron's value. By clamping the gradients, we train the encoder
to put all the information about the variations in the batch into one
output neuron.

\subsection{Invariance Targeting}\label{sec:targetedinvar}

By training with only one transformation at a time, we are encouraging
certain neurons to contain specific information; this is equivariance.
But we also wish to explicitly \emph{discourage} them from having
\emph{other} information; that is, we want them to be invariant to other
transformations. Since our mini-batches of training data consist of only
one transformation per batch, then this goal corresponds to having all
but one of the output neurons of the encoder give the same output for
every image in the batch.

To encourage this property of the DC-IGN, we train all the neurons which
correspond to the inactive transformations with an error gradient equal
to their difference from the mean. It is simplest to think about this
gradient as acting on the set of subvectors $z_{inactive}$ from the
encoder for each input in the batch. Each of these $z_{inactive}$'s
will be pointing to a close-together but not identical point in a
high-dimensional space; the invariance training signal will push them
all closer together. We don't care where they are; the network can
represent the face shown in this batch however it likes. We only care
that the network always represents it as still being the same face, no
matter which way it's facing. This regularizing force needs to be scaled
to be much smaller than the true training signal, otherwise it can
overwhelm the reconstruction goal. Empirically, a factor of $1/100$
works well.

\section{Experiments}\label{experiments}

\begin{figure}[htbp]
\centering
\includegraphics{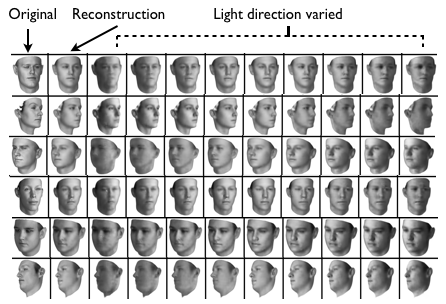}
\caption{\label{fig:manipulating_light}\textbf{Manipulating light.}
Qualitative results showing the generalization capability of the learned
DC-IGN decoder to re-render a single input image under different
lighting conditions. We change the latent $z_{light}$ smoothly leaving
all 199 other latents unchanged.}
\end{figure}

We trained our model on about 12,000 batches of faces generated from a
3D face model obtained from Paysan et al.
(\protect\hyperlink{ref-paysan2009face}{2009}), where each batch
consists of 20 faces with random variations on face identity variables
(shape/texture), pose, or lighting. We used the \emph{rmsprop} (Tieleman
and Hinton \protect\hyperlink{ref-rmsprop}{2012}) learning algorithm
during training and set the meta learning rate equal to $0.0005$, the
momentum decay to $0.1$ and weight decay to $0.01$.

\begin{figure}[htbp]
\centering
\includegraphics{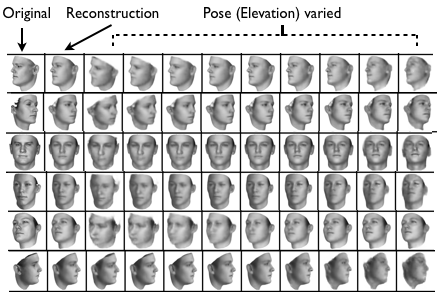}
\caption{\label{fig:manipulating_elevation}\textbf{Manipulating
elevation.} Results showing the ability of the DC-IGN decoder to change
the elevation of the input image. We change the latent $z_{elevation}$
smoothly leaving all 199 other latents unchanged.}
\end{figure}

To ensure that these techniques work on other types of data, we also
trained networks to perform reconstruction on images of widely varied 3D
chairs from many perspectives derived from the Pascal Visual Object
Classes dataset as extracted by Aubry et al. (Mottaghi et al.
\protect\hyperlink{ref-mottaghi2014role}{2014}, Aubry et al.
(\protect\hyperlink{ref-aubry2014seeing}{2014})). This task tests the
ability of the DC-IGN to learn a rendering function for a dataset with
high variation between the elements of the set; the chairs vary from
office chairs to wicker to modern designs, and viewpoints span 360
degrees and two elevations. These networks were trained with the same
methods and parameters as the ones above.

\begin{figure}[htbp]
\centering
\includegraphics{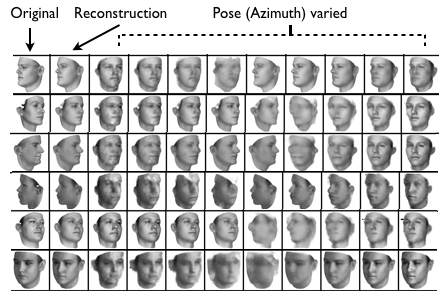}
\caption{\label{fig:manipulating_azimuth}\textbf{Manipulating azimuth
(horizontal angle).} Qualitative results showing the generalization
capability of the learnt DC-IGN decoder to render original static image
with different azimuth (pose) directions. The latent neuron
$z_{azimuth}$ is changed to random values but all other latents are
clamped.}
\end{figure}

\subsection{3D Face Dataset}\label{sec:gen}

The decoder network learns an approximate rendering engine as shown in
Fig.~\ref{fig:manipulating_light}. Given a static test image, the
encoder network produces the latents $Z$ depicting scene variables
such as light, pose, shape etc. Similar to an off-the-shelf rendering
engine, we can independently control these to generate new images with
the decoder. For example, as shown in Fig.~\ref{fig:manipulating_light},
given the original test image, we can vary the lighting of an image by
keeping all the other latents constant and varying $z_{light}$. It is
perhaps surprising that the fully-trained decoder network is able to
function as a 3D rendering engine, and this capability is proof that the
representation learned by the DC-IGN is disentangled.

\begin{figure}[htbp]
\centering
\includegraphics[height=0.95000\textwidth]{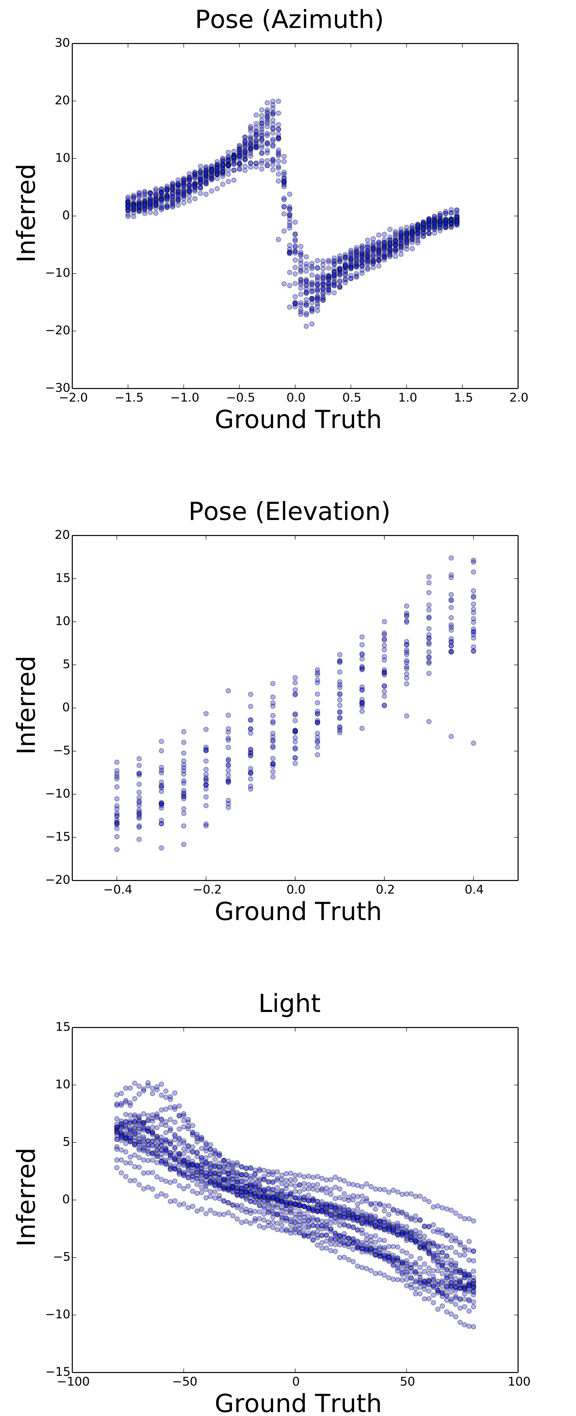}
\caption{\label{fig:gen}\textbf{Generalization of decoder to render
images in novel viewpoints and lighting conditions.} We generated
several datasets by varying light, azimuth and elevation, and tested the
invariance properties of DC-IGN's representation $Z$. We show
quantitative performance on three network configurations as described in
Sec.~\ref{sec:gen}. All DC-IGN encoder networks reasonably predicts
transformations from static test images. Interestingly, as seen in the
first plot, the encoder network seems to have learnt a \emph{switch}
node to deal uniquely with the mirror-symmetric sides of the face.}
\end{figure}

We also quantitatively illustrate the network's ability to represent
pose and light on a smooth linear manifold as shown in
Fig.~\ref{fig:gen}, which directly demonstrates our training algorithm's
ability to disentangle complex transformations. In these plots, the
inferred and ground-truth transformation values are plotted for a random
subset of the test set. Interestingly, as shown in Fig.~\ref{fig:gen},
the encoder network's representation of azimuth has a discontinuity at
$0^\circ$ (facing straight forward).

\begin{figure}[htbp]
\centering
\includegraphics{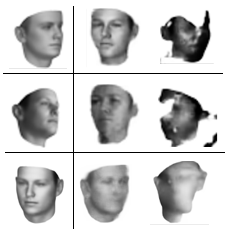}
\caption{\label{fig:pose_and_entangledcomparison}\textbf{Entangled
versus disentangled representations.} \emph{First column:} Original
images. \emph{Second column:} transformed image using DC-IGN.
\emph{Third column:} transformed image using normally-trained network.}
\end{figure}

\subsection{Comparison with Entangled
Representations}\label{comparison-with-entangled-representations}

To explore how much of a difference the DC-IGN training procedure makes,
we compare the novel-view reconstruction performance of networks with
entangled representations (baseline) versus disentangled representations
(DC-IGN). The baseline network is identical in every way to the DC-IGN,
but was trained with SGVB without using our proposed training procedure.
As in Fig.~\ref{fig:manipulating_azimuth}, we feed each network a single
input image, then attempt to use the decoder to re-render this image at
different azimuth angles. To do this, we first must figure out which
latent of the entangled representation most closely corresponds to the
azimuth. This we do rather simply. First, we encode all images in an
azimuth-varied batch using the baseline's encoder. Then we calculate the
variance of each of the latents over this batch. The latent with the
largest variance is then the one most closely associated with the
azimuth of the face, and we will call it $z_{azimuth}$. Once that is
found, the latent $z_{azimuth}$ is varied for both the models to
render a novel view of the face given a single image of that face.
Fig.~\ref{fig:pose_and_entangledcomparison} shows that explicit
disentanglement is critical for novel-view reconstruction.

\subsection{Chair Dataset}\label{chair-dataset}

\begin{figure}[htbp]
\centering
\includegraphics{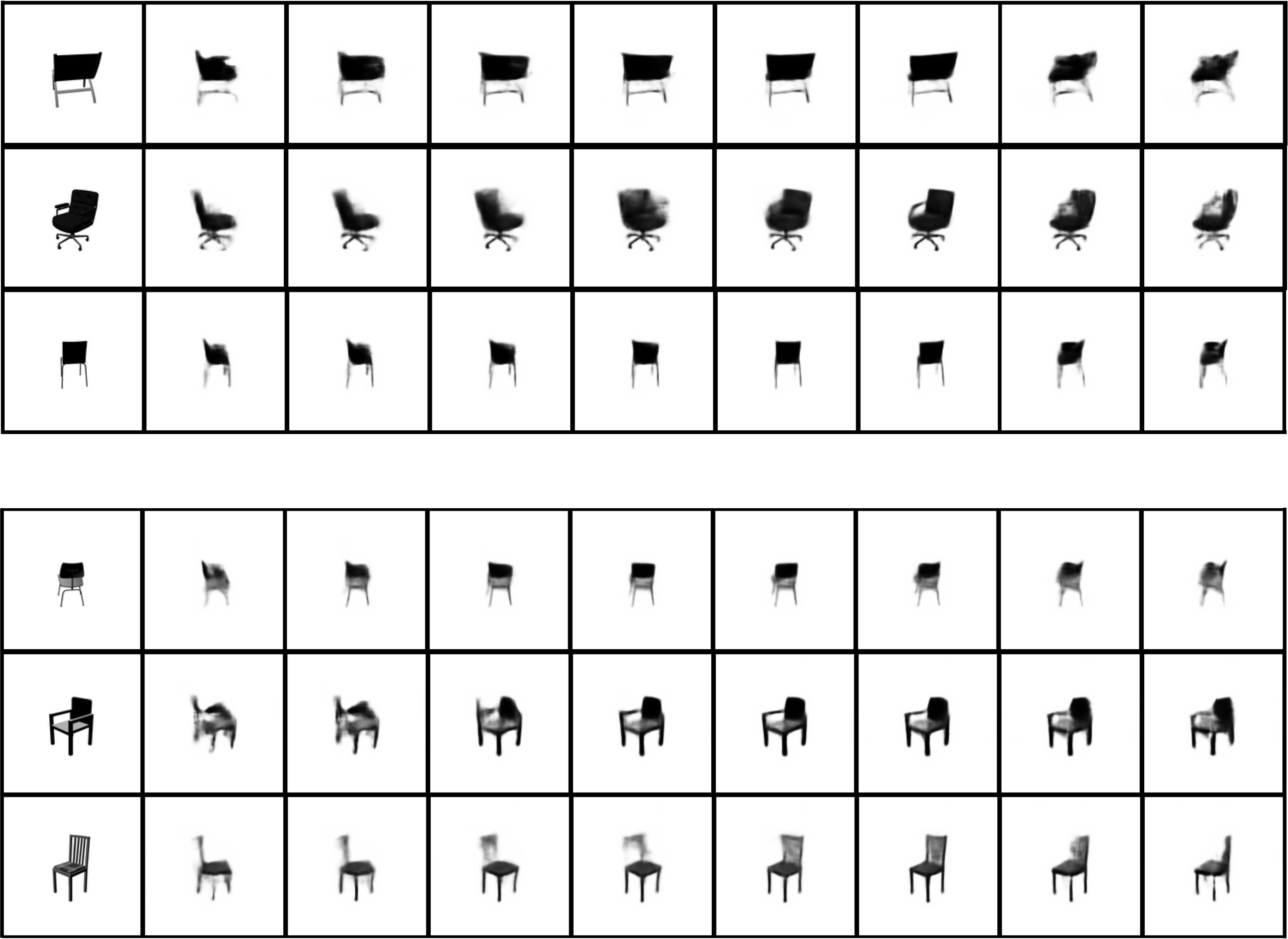}
\caption{\label{fig:chairpose}\textbf{Manipulating rotation:} Each row
was generated by encoding the input image (leftmost) with the encoder,
then changing the value of a single latent and putting this modified
encoding through the decoder. The network has never seen these chairs
before at any orientation. \textbf{Top:} Some positive examples. Note
that the DC-IGN is making a conjecture about any components of the chair
it cannot see; in particular, it guesses that the chair in the top row
has arms, because it can't see that it doesn't. \textbf{Bottom:}
Examples in which the network extrapolates to new viewpoints less
accurately.}
\end{figure}

We performed a similar set of experiments on the 3D chairs dataset
described above. This dataset contains still images rendered from 3D CAD
models of 1357 different chairs, each model skinned with the
photographic texture of the real chair. Each of these models is rendered
in 60 different poses; at each of two elevations, there are 30 images
taken from 360 degrees around the model. We used approximately 1200 of
these chairs in the training set and the remaining 150 in the test set;
as such, the networks had never seen the chairs in the test set from any
angle, so the tests explore the networks' ability to generalize to
arbitrary chairs. We resized the images to $150 \times 150$ pixels and
made them grayscale to match our face dataset.

We trained these networks with the azimuth (flat rotation) of the chair
as a disentangled variable represented by a single node $z_1$; all
other variation between images is undifferentiated and represented by
$z_{[2,200]}$. The DC-IGN network succeeded in achieving a
mean-squared error (MSE) of reconstruction of $2.7722 \times 10^{-4}$
on the test set. Each image has grayscale values in the range $[0,1]$
and is $150 \times 150$ pixels.

In Fig.~\ref{fig:chairpose} we have included examples of the network's
ability to re-render previously-unseen chairs at different angles given
a single image. For some chairs it is able to render fairly smooth
transitions, showing the chair at many intermediate poses, while for
others it seems to only capture a sort of ``keyframes'' representation,
only having distinct outputs for a few angles. Interestingly, the task
of rotating a chair seen only from one angle requires speculation about
unseen components; the chair might have arms, or not; a curved seat or a
flat one; etc.

\chapter{Disentanglement in
Computation}\label{disentanglement-in-computation}

\section{Introduction}\label{introduction-2}

While the learning of representation in fields such as vision and
language have been extensively studied, representations in computation
have only recently begun to be studied.

Instead of thinking about \emph{data}, like images or text,
representations of computation are about representing \emph{procedures},
computation itself. Instead of representing an image, we might represent
a transformation of that image, like rotating every object in it by 90°.
Instead of representing the words in an English phrase, we might
represent a program which translates it into French.

Every neural network can be thought of as a representation of a
computation. The weights and nonlinearities in the networks combine to
transform some input data to some output data; in action they are a
\emph{function} from an input domain to an output domain, and in storage
they represent this function. The famous ImageNet network by Krizhevsky
et al. (\protect\hyperlink{ref-krizhevsky2012imagenet}{2012}), for
example, contains 60 million parameters which, along with their
connectivity, define a function from an input domain of 256x256x3 image
to a 1x1000 distribution over labels.

Applying our desiderata for representations, let us consider the quality
of this 60-million-weight representation for a function which classifies
images.

\begin{enumerate}
\def\labelenumi{\arabic{enumi}.}
\item
  \textbf{Disentangled}: Just as any representation of data should be
  sparse over real transformations, the representation of the
  transformation itself should be sparse. The only clear factorization
  of the computation represented by a feedforward neural network is the
  factorization into layers. Each layer of the network represents a
  large matrix multiplication, and the function computed is the same for
  all inputs. This representation for computation is not at all sparse
  over its inputs, for the entire computation is performed no matter
  what the input is.
\item
  \textbf{Interpretable}: Neural networks are famously hard to
  interpret. Researchers have developed whole classes of techniques for
  analyzing them, which use gradient ascent to visualize specific units
  of the network (Erhan et al.
  \protect\hyperlink{ref-erhan2009visualizing}{2009}), occlusions of the
  input to analyze significance (Zeiler and Fergus
  \protect\hyperlink{ref-zeiler2014visualizing}{2014}), or inverting
  their functions to visualize their information preservation (Mahendran
  and Vedaldi \protect\hyperlink{ref-mahendran2014understanding}{2014}).
  These techniques speak to the deeply uninterpretable nature of neural
  representations of computation.
\item
  \textbf{Performant}: Deep neural networks currently hold the accuracy
  records in almost every large-N dataset of image recognition, object
  localization, and phoneme recognition benchmark. In particular, this
  network set the record for ImageNet performance with an error rate
  more than 40\% lower than any other entry.
\item
  \textbf{Reusable}: While substantial reuse of pretrained copies of
  this network has been made, such reuse is by no means simple.
  Typically the top half of the network is completely removed and
  another is trained (quite expensively) in its place; in other use
  cases the lower-level features generated by the first few layers of
  the network have been used directly as an embedding of the input
  space, with very mixed results. Compared with a more modular design,
  which might have separate components for localizing salient objects
  and determining various salient information about them (size, color,
  animacy, shape, context) this representation is quite hard to reuse.
\item
  \textbf{Compact}: This model contains 60 million parameters. It
  occupies hundreds of megabytes on disk when compressed. While those
  numbers sound large, it is not immediately clear if this is very large
  or very compact for a model which contains all necessary information
  for determining the contents of arbitrary images.
\end{enumerate}

While this model performs extremely well, and might (in bad lighting,
with the right Instagram filter) be considered compact, it is very far
from ideal in disentanglement, interpretability, and reusability. Just
by disentangling the computation in this model, factorizing it into
modules, its interpretability and reusability would be hugely improved.

If, for example, this network were disentangled by having a module which
determined whether a scene was indoors or outdoors and a separate
classifier for each of those cases, we would gain several advantages:

\begin{itemize}
\tightlist
\item
  The indoor/outdoor classifier would be immediately comprehensible.
\item
  The indoor/outdoor classifier could be reused in other tasks.
\item
  The location-specific object classifiers could be more easily
  interpreted (e.g.~you would be very surprised if the indoor classifier
  predicted a train, or a gorilla).
\item
  The location-specific object classifiers would generate intermediate
  features which were more diagnostic for other tasks in their given
  location.
\end{itemize}

To make an unfair comparison, let's use our desiderata to consider the
quality of a Python representation of the computation of the FizzBuzz
problem:

\hypertarget{lst:fizzbuzz}{\label{lst:fizzbuzz}}
\begin{Shaded}
\begin{Highlighting}[]
\KeywordTok{def} \NormalTok{divisible_by_five(n):}
    \ControlFlowTok{return} \NormalTok{n }\OperatorTok{%} \DecValTok{5} \OperatorTok{==} \DecValTok{0}

\KeywordTok{def} \NormalTok{divisible_by_three(n):}
    \ControlFlowTok{return} \NormalTok{n }\OperatorTok{%} \DecValTok{3} \OperatorTok{==} \DecValTok{0}

\KeywordTok{def} \NormalTok{fizzbuzz(n):}
    \NormalTok{result }\OperatorTok{=} \StringTok{''}
    \ControlFlowTok{if} \NormalTok{divisible_by_three(n):}
        \NormalTok{result }\OperatorTok{+=} \StringTok{'fizz'}
    \ControlFlowTok{if} \NormalTok{divisible_by_five(n):}
        \NormalTok{result }\OperatorTok{+=} \StringTok{'buzz'}
    \ControlFlowTok{return} \NormalTok{result}

\KeywordTok{def} \NormalTok{fizzbuzz_string(length):}
    \NormalTok{result_list }\OperatorTok{=} \BuiltInTok{map}\NormalTok{(fizzbuzz, }\BuiltInTok{range}\NormalTok{(}\DecValTok{1}\NormalTok{, length }\OperatorTok{+} \DecValTok{1}\NormalTok{))}
    \ControlFlowTok{return} \StringTok{'}\CharTok{\textbackslash{}n}\StringTok{'}\NormalTok{.join(result_list)}

\BuiltInTok{print}\NormalTok{(fizzbuzz_string(}\DecValTok{100}\NormalTok{))}
\end{Highlighting}
\end{Shaded}

\begin{enumerate}
\def\labelenumi{\arabic{enumi}.}
\item
  \textbf{Disentangled}: This computation has been factorized into a
  number of distinct subcomponents, each of which is very small and can
  be used in multiple places. They do not depend on the state of the
  overall program, and have very low-dimensional and clearly-defined
  inputs and outputs. Simple operators such as \texttt{+=} or
  \texttt{\%} are composed into larger ones, and the contribution from
  each is very clear.
\item
  \textbf{Interpretable}: This representation can be easily read by
  anyone who knows how to program, and most of it could be understood
  even by people who don't.
\item
  \textbf{Performant}: While this code will make no errors on the task,
  this is not a meaningful question on a toy task.
\item
  \textbf{Reusable}: Individual components of this code represent
  functions which could be used elsewhere or for variants of this task.
  It would be trivial to use \texttt{divisible\_by\_three} anywhere else
  its functionality is needed, and the other functions can similarly be
  reused to generate FizzBuzz solutions of any length.
\item
  \textbf{Compact}: This representation occupies 370 bytes.
\end{enumerate}

Our ideal representation of a computation would share the learnability
and performance on hard problems of the deep network without giving up
the goals of disentanglement and reuse quite so completely as the deep
network does.

\subsection{Catastrophic forgetting}\label{catastrophic-forgetting}

One of the clearest demonstrations of the weakness of highly-entangled
neural network representations of computation is catastrophic
forgetting.

With the recent success of deep learning methods in many fields, efforts
have been made to apply deep learning techniques to multitask learning
problems. Deep learning is at the deepest level a method for
hierarchically extracting good representations from complex data, with
the higher levels of a network capturing increasingly abstract
representations of the data. As such, deep learning seems naively to be
a promising direction for multitask learning; abstract representations
of the data should be useful for many related tasks, and the network
should be able to simply not use any which are not helpful.

This theory has been borne out for simple, highly coupled tasks such as
evaluating sentiment of reviews for different categories of products
(Glorot, Bordes, and Bengio
\protect\hyperlink{ref-glorot2011domain}{2011}). A more wide-ranging
survey of deep learning methods for transfer and multitask learning
shows that some classes of models are able to improve their performance
on the original, clean dataset after being shown perturbed or distorted
versions of the same data (Bengio
\protect\hyperlink{ref-bengio2012deep}{2012}).

However, even small changes in the task result in substantial changes to
the optimal features, especially at high levels of the network (Yosinski
et al. \protect\hyperlink{ref-yosinski2014transferable}{2014}). This can
lead to \emph{catastrophic forgetting}, in which the network
``unlearns'' one task as it trains on another one. A recent set of
experiments (Goodfellow et al.
\protect\hyperlink{ref-goodfellow2013empirical}{2013}) detail the
tradeoff curve for performance on one task versus performance on the
other task for both similar and dissimilar tasks. They show that for
networks trained on two tasks, improvement on one task comes at a cost
to performance on another.

This occurs because of the highly entangled nature of the computation
carried out by these networks. When the network which is able to solve
two different tasks is retrained on just one, it gradually mutates the
calculations which are necessary in both of the tasks, until eventually
it has repurposed them entirely for the use of the first task. What this
system needs is a clear separation of concerns. If some functional
elements of the network were used only for one task, those elements
would be safe to mutate at a high rate during training on that task.
Similarly, those elements which were used across many tasks could change
only very gradually, ensuring that even if one task is neglected for an
extended period, the components it uses won't have diverged too greatly
from their original state.

In more specific terms, the problem is that each weight in the network
receives gradients of a similar magnitude when training on either task.
And with no parameters ``reserved'' for a specific task, that task is
quickly forgotten.

\section{Related Work}\label{related-work-1}

Until recently, work in this domain has largely centered around either
a) learning programs in a fixed representation language, or b) jointly
learning a program and its representation in e.g.~a neural network, but
with little attention focus on the representation itself. In particular,
Liang et al. (Liang, Jordan, and Klein
\protect\hyperlink{ref-liang2010learning}{2010}) propose to learn
programs via Bayesian inference with a grammar over a hierarchical
structure. Zaremba et al.
(\protect\hyperlink{ref-zaremba2014learning}{2014}) use an LSTM
(Hochreiter and Schmidhuber
\protect\hyperlink{ref-hochreiter1997long}{1997}) to predict the output
of simple programs written in Python; their effectiveness is remarkable,
but the induced representation is so poor that the authors comment, ``We
do not know how heavily our model relies on memorization and how far the
learned algorithm is from the actual, correct algorithm.''

A classic model that attempts to disentangle computation is the mixture
of experts (Jacobs, Jordan, and Barto
\protect\hyperlink{ref-jacobs1991task}{1991}). However, as originally
described this model was not especially successful at learning distinct
functions for each expert; this led to a modification of the design
which used sampling instead of weighting using the gating values (Jacobs
et al. \protect\hyperlink{ref-jacobs1991adaptive}{1991}). This modified
design resulted in nicely decoupled functions, but was much harder to
train. Addressing this problem was a core inspiration for my work.

In the last year, work on learning structured representations of
computation has become a popular topic. (Neelakantan, Le, and Sutskever
\protect\hyperlink{ref-neelakantan2015neural}{2015}) augment a neural
network with a small set of hard-coded external operations which the
network learns to use in multistep programs. (Reed and Freitas
\protect\hyperlink{ref-reed2015neural}{2015}) propose a very general
model which similarly can use external programs, but with the addition
of a call stack; however, this model requires strong supervision to
train explicitly with the correct program trace, and as such is learning
to recreate an existing program representation. (Zaremba et al.
\protect\hyperlink{ref-zaremba2015learning}{2015}) use an external
memory with pointers to learn routines for interacting with external
data. (Graves, Wayne, and Danihelka
\protect\hyperlink{ref-graves2014neural}{2014}) perform complex
operations on sequences such as sorting or repeatedly copying by using a
differentiable content-based addressing mechanism to read and write to
an external memory.

\section{Controller-function
networks}\label{controller-function-networks}

\begin{figure}[htbp]
\centering
\includegraphics{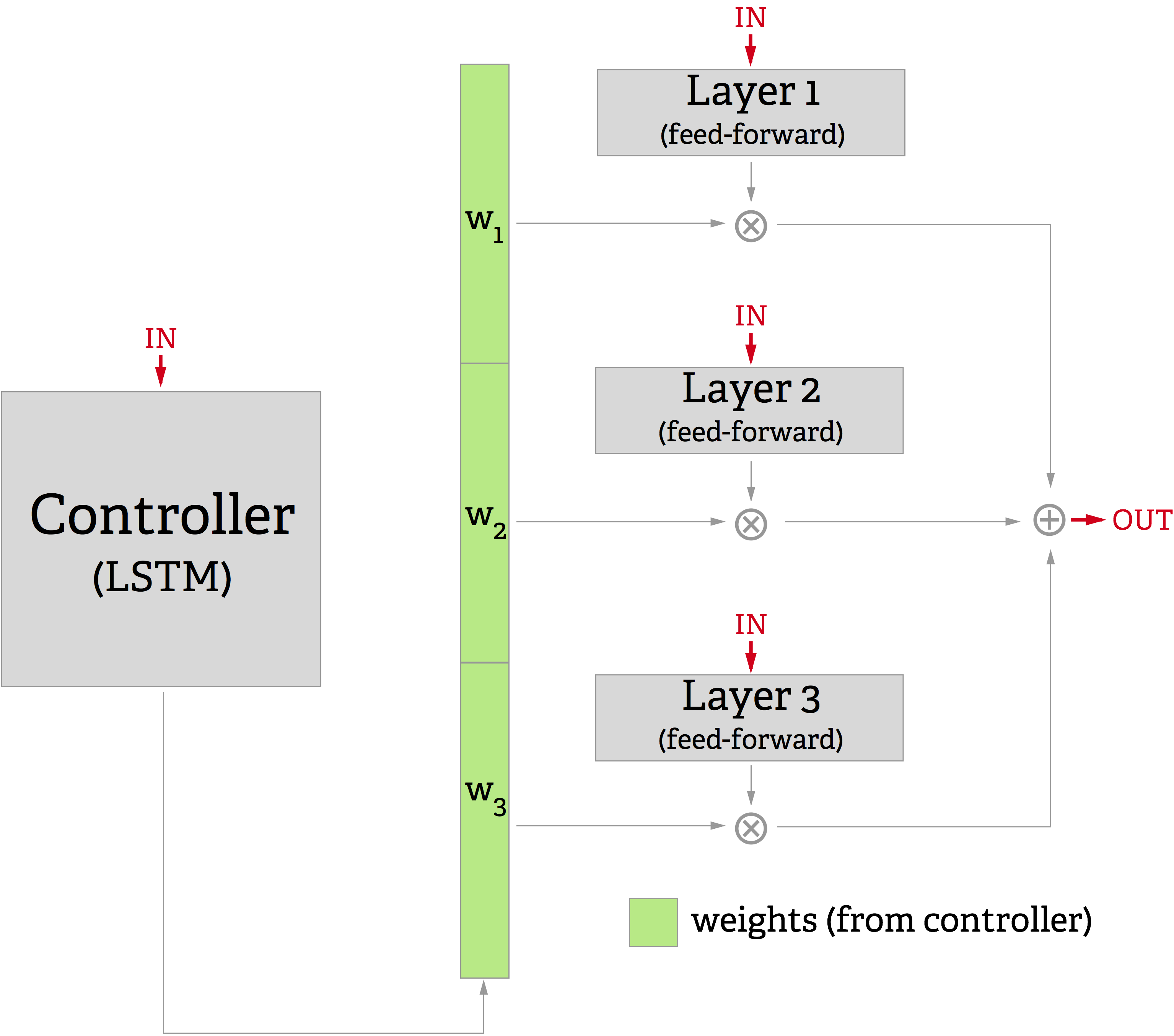}
\caption{\label{fig:controller_network}\textbf{The controller and layers
of the controller-function network (CFN).} The controller provides
weights on each layer as a function of the data. This shows three
layers, but there can be many more.}
\end{figure}

The proposed model, the controller-function network (CFN) generates an
output for a particular timestep via the following steps (shown in
Fig.~\ref{fig:controller_network}):

\begin{enumerate}
\def\labelenumi{\arabic{enumi}.}
\tightlist
\item
  The input tensor is fed into the controller
\item
  The controller decides which layers are most appropriate for
  processing this input
\item
  The controller outputs a weighting vector reflecting how much output
  it wants from each of the layers
\item
  The input tensor is fed into each layer (in parallel)
\item
  The outputs from each layer are multiplied by their respective weights
  from the controller
\item
  The weighted outputs from all the layers are summed together and
  output. This is the output of the whole network for this timestep.
\end{enumerate}

Essentially the idea is that at each timestep, the controller examines
the input that it gets, then produces a distribution over the activities
of the various ``functions'' (single-layer NNs) which would best deal
with this input. Since the controller is an LSTM, it can store
information about the inputs it has received before, meaning that in a
time series or language setting it can make weighting decisions
contextually.

Each of the ``function'' layers is a single-layer network with a PReLU
activation function (He et al.
\protect\hyperlink{ref-he2015delving}{2015}). The input and output
dimension of these functions is always the same, and corresponds to the
desired output dimension of the network as a whole.

As this model is differentiable throughout, it can be trained with the
standard backpropagation through time (BPTT) algorithm for stochastic
gradient descent.

By setting weights over each of the layers in the network, the
controller scales not only the output of each layer, but also the error
gradient that it receives. This means that in a given timestep, the
layers which have very low weights on their output will be nearly
unchanged by the learning process. That is, functions which are not used
are not forgotten.

In an ordinary feedforward neural network, the only way for the network
to prevent learning in a particular node is for it to learn connection
strengths very near zero for that node. This takes many training
examples, and functionally removes that node from the computation graph.

This system, by comparison, can decide that a set of nodes is or is not
relevant on an input-by-input basis.

\subsection{Relationship to mixture of
experts}\label{relationship-to-mixture-of-experts}

This architecture is closely related to the mixture of experts model
proposed by Jacobs et al.
(\protect\hyperlink{ref-jacobs1991task}{1991}), in which several
different task-specific ``expert'' networks each contribute in linear
combination to the output of the overall network.

However, this model has two key differences from the mixture of experts:

\begin{enumerate}
\def\labelenumi{\arabic{enumi}.}
\tightlist
\item
  \textbf{The gating network is an LSTM.} This means that the gating
  network (or controller, in my terminology) can easily learn fixed
  sequential procedures for certain types of input. This allow the model
  to be iterated for several steps, composing its operations into more
  complex ones. See Sec.~\ref{sec:multistep} for a description of this
  usage.
\item
  \textbf{The training results in decoupled functions.} I employ a novel
  continuation method for training the CFN that allows for easy
  training, yet results in a final representation which uses only one
  ``expert'' at a time with no overlap.
\end{enumerate}

\subsection{Hard and soft decisions}\label{hard-and-soft-decisions}

Training neural models which make ``hard'' decisions can be quite
challenging; in the general case, such models must be trained by
REINFORCE-like gradient estimation methods (Williams
\protect\hyperlink{ref-williams1992simple}{1992}). Yet under many
circumstances, such hard decisions are necessary for computational
considerations; in fully-differentiable models such as the NTM (Graves,
Wayne, and Danihelka \protect\hyperlink{ref-graves2014neural}{2014}) or
end-to-end memory networks (Sukhbaatar et al.
\protect\hyperlink{ref-sukhbaatar2015end}{2015}), the computational
complexity of a single evaluation increases linearly with the size of
the memory. These ``soft'' decisions involve considering every possible
option.

In more complex computational tasks, such as those faced by (Reed and
Freitas \protect\hyperlink{ref-reed2015neural}{2015}), there may be a
large number of steps before any result is produced by the model, and
each step can require a discrete action (move the pointer either left or
right; move the model either up or down). Such models naïvely have
branching which is exponential of the form $O(k^t)$, where $k$ is
the number of options at each timestep, and $t$ is the number of
timesteps before producing an output. Using a REINFORCE algorithm to
estimate the true gradient is possible, but slow and unreliable (Zaremba
and Sutskever \protect\hyperlink{ref-zaremba2015reinforcement}{2015}).
This branching factor is what led (Reed and Freitas
\protect\hyperlink{ref-reed2015neural}{2015}) to adopt their
strongly-supervised training technique.

A straightforward (if inelegant) solution is to composite the outcome
from all of these branches at the end of each timestep. For example, a
pointer could be modeled as interpolating between two memory cells
instead of having a discrete location. Then when the controller produces
the distribution of actions ``left 0.7, right 0.3'', the model can move
the pointer left by 0.7 instead of sampling from $Bernoulli(0.7)$.

While such techniques, make the learning process tractable when
available, they result in much more highly entangled representations
(e.g.~reading from a every location in a memory at once). Furthermore,
they must always incur a complexity cost linear in the number of
options, just as the memory models have cost linear in the number of
options of memory locations to read from.

In especially challenging environments, this solution is not available.
For example, in classic reinforcement learning tasks, the agent may only
be in a situation once, and it cannot 70\% fly to Germany or 20\% accept
a PhD position.

The CFN exists in the space of models for which this soft-decision
solution is available. While in the ideal case we would like to select
exactly one function to use at each timestep, this problem is quite
difficult to optimize, for early in training the functions are not yet
differentiated. By contrast, the soft-decision version which uses a
weighted sum of the outputs of each function learns quite quickly.
However, the solutions produced by this weighted sum training are highly
entangled and always involve a linear combination of all the functions,
with no clear differentiation.

From scratch, we can either train a system that works, or a system that
has good representations. What we need is a way to go from a working
solution to a good solution.

\subsection{Continuation methods}\label{sec:continuation}

Continuation methods are a widely-used technique for approaching
difficult optimization problems.

\begin{quote}
In optimization by continuation, a transformation of the nonconvex
function to an easy-to-minimize function is considered. The method then
progressively converts the easy problem back to the original function,
while following the path of the minimizer. (Mobahi and Fisher III
\protect\hyperlink{ref-mobahi2015theoretical}{2015})
\end{quote}

As described in (Mobahi and Fisher III
\protect\hyperlink{ref-mobahi2015theoretical}{2015}), continuations
include ideas as ubiquitous as curriculum learning or deterministic
annealing, and that paper provides an extensive list of examples. In the
quest for good solutions to hard-decision problems, continuation methods
are a natural tool.

\subsection{Training with noisy
decisions}\label{training-with-noisy-decisions}

In order to construct a continuation between soft and hard decisions,
the CFN combines two tools: weight sharpening and noise.

Weight sharpening is a technique used by (Graves, Wayne, and Danihelka
\protect\hyperlink{ref-graves2014neural}{2014}), which works by taking a
distribution vector of weights $w \in [0,1]^n$, and a sharpening
parameter $\gamma \ge 1$ and transforming $w$ as follows:

\[w_i' = \frac{w_i^{\gamma}}{\sum_j w_j^{\gamma}}\]

By taking this $[0,1]^n$ vector to an exponent, sharpening increases
the relative differences between the weights in $w$. Renormalizing
makes $w$ a distribution once again, but now it has been stretched;
large values are larger, i.e.~the modes have higher probability. In the
CFN, I take one further step: adding noise.

\[w_i' = \frac{\big(w_i + \mathcal{N}(0, \sigma^2)\big)^{\gamma}}{\sum_j w_j^{\gamma}}\]

During the training of the CFN, sharpening is applied to the vector of
weights produced by the controller, and the sharpening parameter
$\gamma$ is gradually increased on a schedule. By itself, this would
not transform the outputs of the controller, as it can simply learn the
inverse function to continue to produce the same output. However, the
addition of noise before sharpening makes similar weights highly
unstable. For example, if the network \emph{intended} to produce a
weighting of $[0.5, 0.5]$, noise would interfere:

\[\frac{[0.49, 0.51]^{100}}{(0.49^{100} + 0.51^{100})} = [0.018, 0.982]\]

At the end of training, this forces the CFN to either make a hard
decision or face massive uncertainty in its output. By slowly increasing
the sharpening parameter on a schedule, the controller can gradually
learn to make harder and harder decisions. In practice this method works
very well, resulting in perfectly binary decisions at the end of
training and correct factorization of the primitives, each into its own
function layer.

\section{Experiments}\label{experiments-1}

In order to carefully test the ability of various techniques to
correctly factorize several presented problems, I constructed a simple
dataset of vector functions, inputs, and outputs. These functions are
detailed in Tbl.~\ref{tbl:primitives}. In the following experiments,
these functions are applied to random input vectors in $[0,1]^{10}$.

Since the inputs to all of these functions are indistinguishable,
without any extra information it would be impossible for any system to
achieve results better than averaging the output of all these functions.
Therefore, along with the input vector, all systems receive a one-hot
vector containing the index of the primitive to compute. Each system
must learn to interpret this information in its own way. In the CFN,
this metadata is passed only to the controller, which forces it to use
different functions for different inputs.

While this is on the surface a supervised learning task (given some
input, produce exactly this output), the much more interesting
interpretation of the task is unsupervised. The true goal of this task
is to learn a \emph{representation of computation} which mirrors the
true factorization of the functions which generated this data. If we are
interested in disentangled representations, we should look for systems
which activate very distinct units for each of these separate
computations.

\begin{longtable}[c]{@{}lll@{}}
\caption{\label{tbl:primitives}\textbf{Primitive functions.} The true
test of a learned model is how distinctly the model manages to represent
these functions, not the exact error number. Outputs shown for the input
vector {[}1 2 3 4 5 6 7 8{]}. }\tabularnewline
\toprule
\begin{minipage}[b]{0.14\columnwidth}\raggedright\strut
Operation
\strut\end{minipage} &
\begin{minipage}[b]{0.55\columnwidth}\raggedright\strut
Description
\strut\end{minipage} &
\begin{minipage}[b]{0.22\columnwidth}\raggedright\strut
Output
\strut\end{minipage}\tabularnewline
\midrule
\endfirsthead
\toprule
\begin{minipage}[b]{0.14\columnwidth}\raggedright\strut
Operation
\strut\end{minipage} &
\begin{minipage}[b]{0.55\columnwidth}\raggedright\strut
Description
\strut\end{minipage} &
\begin{minipage}[b]{0.22\columnwidth}\raggedright\strut
Output
\strut\end{minipage}\tabularnewline
\midrule
\endhead
\begin{minipage}[t]{0.14\columnwidth}\raggedright\strut
rotate
\strut\end{minipage} &
\begin{minipage}[t]{0.55\columnwidth}\raggedright\strut
Move each element of the vector right one slot. Move the last component
to the first position.
\strut\end{minipage} &
\begin{minipage}[t]{0.22\columnwidth}\raggedright\strut
{[}8 1 2 3 4 5 6 7{]}
\strut\end{minipage}\tabularnewline
\begin{minipage}[t]{0.14\columnwidth}\raggedright\strut
add-a-b
\strut\end{minipage} &
\begin{minipage}[t]{0.55\columnwidth}\raggedright\strut
Add the second half of the vector to the first.
\strut\end{minipage} &
\begin{minipage}[t]{0.22\columnwidth}\raggedright\strut
{[}6 8 10 12 5 6 7 8{]}
\strut\end{minipage}\tabularnewline
\begin{minipage}[t]{0.14\columnwidth}\raggedright\strut
rot-a
\strut\end{minipage} &
\begin{minipage}[t]{0.55\columnwidth}\raggedright\strut
Rotate only the first half of the vector.
\strut\end{minipage} &
\begin{minipage}[t]{0.22\columnwidth}\raggedright\strut
{[}4 1 2 3 5 6 7 8{]}
\strut\end{minipage}\tabularnewline
\begin{minipage}[t]{0.14\columnwidth}\raggedright\strut
switch
\strut\end{minipage} &
\begin{minipage}[t]{0.55\columnwidth}\raggedright\strut
Switch the positions of the first and second halves of the vector.
\strut\end{minipage} &
\begin{minipage}[t]{0.22\columnwidth}\raggedright\strut
{[}5 6 7 8 1 2 3 4{]}
\strut\end{minipage}\tabularnewline
\begin{minipage}[t]{0.14\columnwidth}\raggedright\strut
zero
\strut\end{minipage} &
\begin{minipage}[t]{0.55\columnwidth}\raggedright\strut
Return the zero vector.
\strut\end{minipage} &
\begin{minipage}[t]{0.22\columnwidth}\raggedright\strut
{[}0 0 0 0 0 0 0 0{]}
\strut\end{minipage}\tabularnewline
\begin{minipage}[t]{0.14\columnwidth}\raggedright\strut
zero-a
\strut\end{minipage} &
\begin{minipage}[t]{0.55\columnwidth}\raggedright\strut
Zero only the first half of the vector.
\strut\end{minipage} &
\begin{minipage}[t]{0.22\columnwidth}\raggedright\strut
{[}0 0 0 0 5 6 7 8{]}
\strut\end{minipage}\tabularnewline
\begin{minipage}[t]{0.14\columnwidth}\raggedright\strut
add-one
\strut\end{minipage} &
\begin{minipage}[t]{0.55\columnwidth}\raggedright\strut
Add 1 to the vector.
\strut\end{minipage} &
\begin{minipage}[t]{0.22\columnwidth}\raggedright\strut
{[}2 3 4 5 6 7 8 9{]}
\strut\end{minipage}\tabularnewline
\begin{minipage}[t]{0.14\columnwidth}\raggedright\strut
swap-first
\strut\end{minipage} &
\begin{minipage}[t]{0.55\columnwidth}\raggedright\strut
Swap the first two elements of the vector.
\strut\end{minipage} &
\begin{minipage}[t]{0.22\columnwidth}\raggedright\strut
{[}2 1 3 4 5 6 7 8{]}
\strut\end{minipage}\tabularnewline
\bottomrule
\end{longtable}

\subsection{Disentanglement of
functions}\label{disentanglement-of-functions}

\begin{figure}[htbp]
\centering
\includegraphics{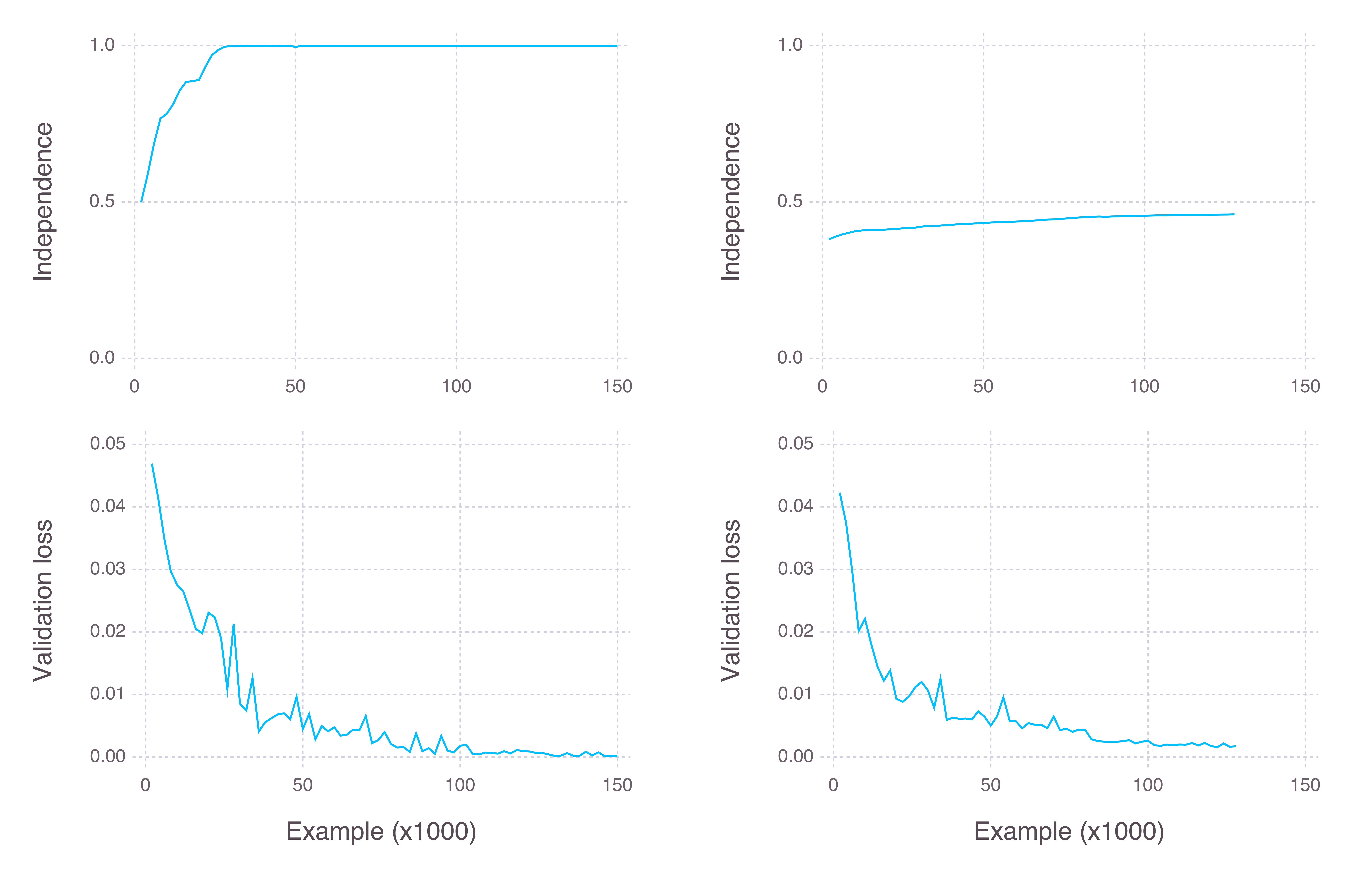}
\caption{\label{fig:loss_entanglement}\textbf{Disentanglement and
validation loss} plotted over the course of training. Disentanglement,
or \emph{independence}, is measured by the L2 norm of the weight vector
over the functions. In this measure, 0.35 is totally entangled, with
every function accorded equal weight for every input, and 1.0 is totally
disentangled, with precisely one function used for each input.
\textbf{Left:} with sharpening and noise. \textbf{Right:} without
sharpening and noise.}
\end{figure}

In order to directly test how disentangled the CFN's representations
are, I analyzed the weights given to each function in the network
throughout the training process. In the ideal case, the distribution
would be entirely concentrated on one function at a time; this would
indicate that the network has perfectly decoupled their functions. Since
no two functions are the same, and they each have the same input domain,
no one function layer can correctly compute two of them.

The results of this analysis are presented in
Fig.~\ref{fig:loss_entanglement}. By using the continuation method
described in Sec.~\ref{sec:continuation}, the CFN is able to very
rapidly learn a disentangled representation of the functions in the data
with no penalty to performance. By comparison, a network of the same
architecture trained without the noise and sharpening technique can also
produce the same output, but its representation of the computation is
very highly entangled.

\subsection{Catastrophic forgetting}\label{catastrophic-forgetting-1}

\begin{figure}[htbp]
\centering
\includegraphics{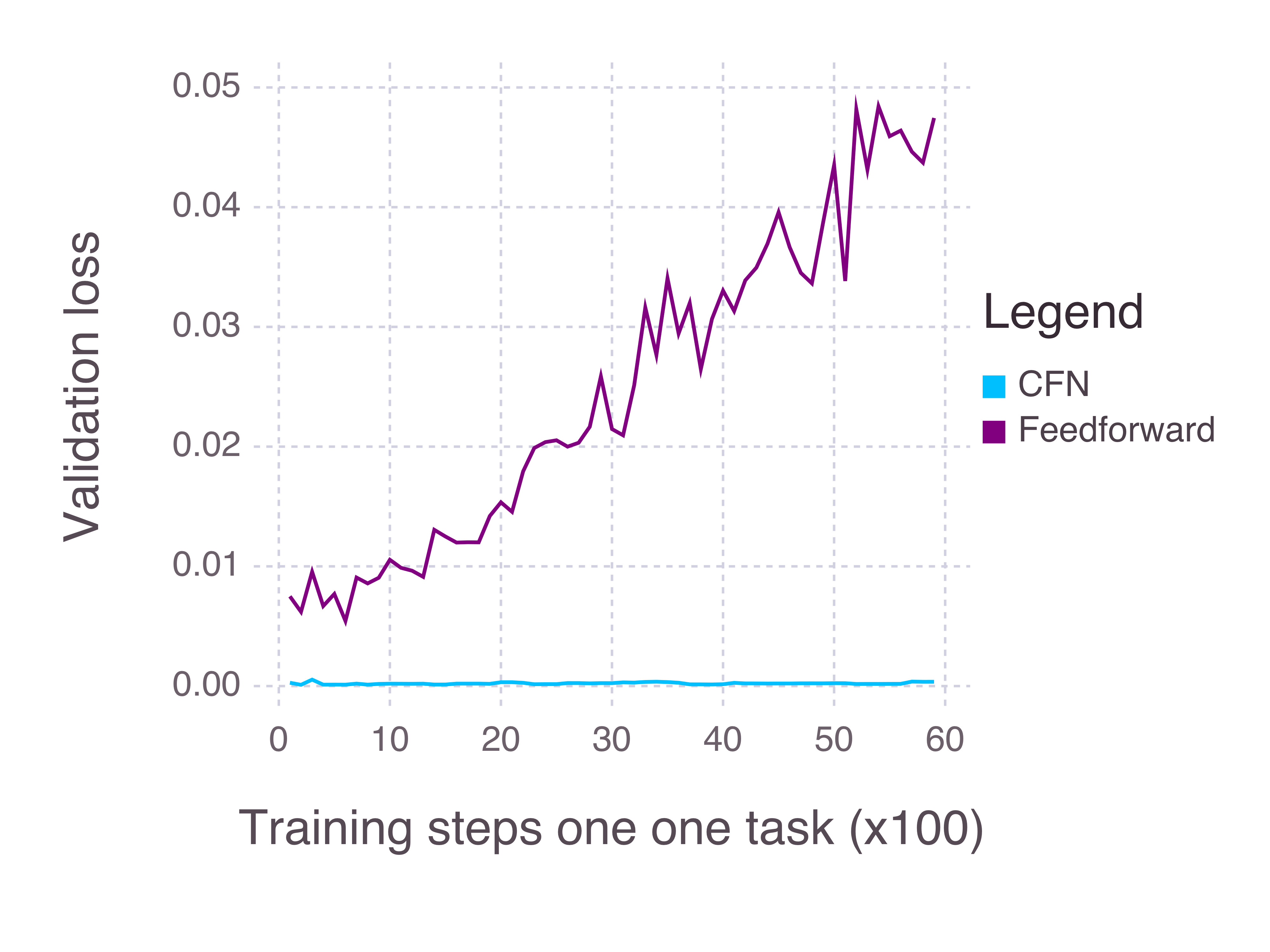}
\caption{\label{fig:forgetting}\textbf{Forgetting when trained on one
task.} When a traditional feedforward network, which previously trained
on several tasks, is trained exclusively on one, it forgets how to
perform the others. The controller-function network is practically
immune to forgetting. In this figure, we see each network trained
exclusively on one of several tasks it is able to do. The loss that is
shown is the average L2 error attained on all of the \emph{other} tasks
as this network retrains.}
\end{figure}

To test the CFN's resistance to the forgetting problems which have
plagued deep methods, I trained a controller-function network and a
feedforward network to convergence on the full dataset, including all
eight vector functions. The feedforward network was densely connected,
with three linear layers of dimension 18, 100, and 10, with PReLu
non-saturating activations in between.

After training each of these networks on the full dataset, I then set
them to training on a dataset consisting of data from only one of the
primitive functions. Both networks were retrained with the same learning
rate and other hyperparameters. Periodically, I evaluated both networks'
performance against a validation set consisting of data generated by all
of the \emph{other} functions. As depicted in Fig.~\ref{fig:forgetting},
the feedforward neural network experienced increasing loss over the
course of training. These results are typical of neural methods. By
contrast, the controller-function network has practically no forgetting
behavior at all; the controller is assigning near-zero weights to all
functions except the correct one, and as a result they receive gradients
very near zero and do not noticeably update.

This result is especially compelling given the difference in parameter
dimensionality of these two models; while the feedforward network has
13013 parameters, the CFN has better performance and better resistance
to forgetting with only 2176. Though feedforward models with fewer
parameters have worse forgetting behavior, the structure of the CFN
representation allows for a very good memory.

\chapter{Discussion}\label{discussion}

\section{DC-IGN}\label{dc-ign}

We have shown that it is possible to train a deep convolutional inverse
graphics network with a fairly disentangled, interpretable graphics code
layer representation from static images. By utilizing a deep convolution
and de-convolution architecture within a variational autoencoder
formulation, our model can be trained end-to-end using back-propagation
on the stochastic variational objective function (Kingma and Welling
\protect\hyperlink{ref-kingma2013auto}{2013}). We proposed a training
procedure to force the network to learn disentangled and interpretable
representations. Using 3D face and chair analysis as a working example,
we have demonstrated the invariant and equivariant characteristics of
the learned representations.

Such a representation is powerful because it teases apart the true
generating factors for images. Unlike a traditional deep representation,
the representation generated by the DC-IGN separates the innate
properties of an object from the results of its particular lighting and
position. This brings us ever so slightly closer to a truly human-like
understanding of 3D scenes, in which we use our knowledge of the
structure of the world to correctly interpret the contributions to an
image from depth, lighting, deformation, occlusion, and so much more. It
is essential that our representations have this structure, as it allows
incredible feats of imagination and prediction that current machine
learning systems cannot even approach.

\subsection{Future work}\label{future-work}

To scale our approach to handle more complex scenes, it will likely be
important to experiment with deeper architectures in order to handle
large number of object categories within a single network architecture.
It is also very appealing to design a spatio-temporal based
convolutional architecture to utilize motion in order to handle
complicated object transformations. Importance-weighted autoencoders
(Burda, Grosse, and Salakhutdinov
\protect\hyperlink{ref-burda2015importance}{2015}) might be better able
to capture more complex probabilistic structure in scenes. Furthermore,
the current formulation of SGVB is restricted to continuous latent
variables. However, real-world visual scenes contain unknown number of
objects that move in and out of frame. Therefore, it might be necessary
to extend this formulation to handle discrete distributions (Kulkarni,
Saeedi, and Gershman
\protect\hyperlink{ref-kulkarni2014variational}{2014}) or extend the
model to a recurrent setting. The decoder network in our model can also
be replaced by a domain-specific decoder (Nair, Susskind, and Hinton
\protect\hyperlink{ref-nair2008analysis}{2008}) for fine-grained
model-based inference.

In scenes with greater multimodal structure, perhaps in the future it
may be possible to impose similar structure on the representation of a
Deep Convolutional Generative Adversarial Network (Radford, Metz, and
Chintala \protect\hyperlink{ref-radford2015unsupervised}{2015}). These
models seem to have the ability to capture much more complicated
distributional structure than the simple Gaussian of the variational
autoencoder.

\section{Controller-function
networks}\label{controller-function-networks-1}

With the design of controller-function networks, I have provided a
modern take on the mixture of experts model and shown that it is
possible learn highly disentangled representations of computation.
Furthermore, I have shown that at least in some cases, doing so imposes
\emph{no performance penalty} on the system. These experiments have also
demonstrated that disentangled representations of computation are much
more resistant to the perennial problem of catastrophic forgetting,
which has plagued neural methods since their inception.

\subsection{Future work}\label{future-work-1}

\subsubsection{Multi-step variant}\label{sec:multistep}

\begin{figure}[htbp]
\centering
\includegraphics{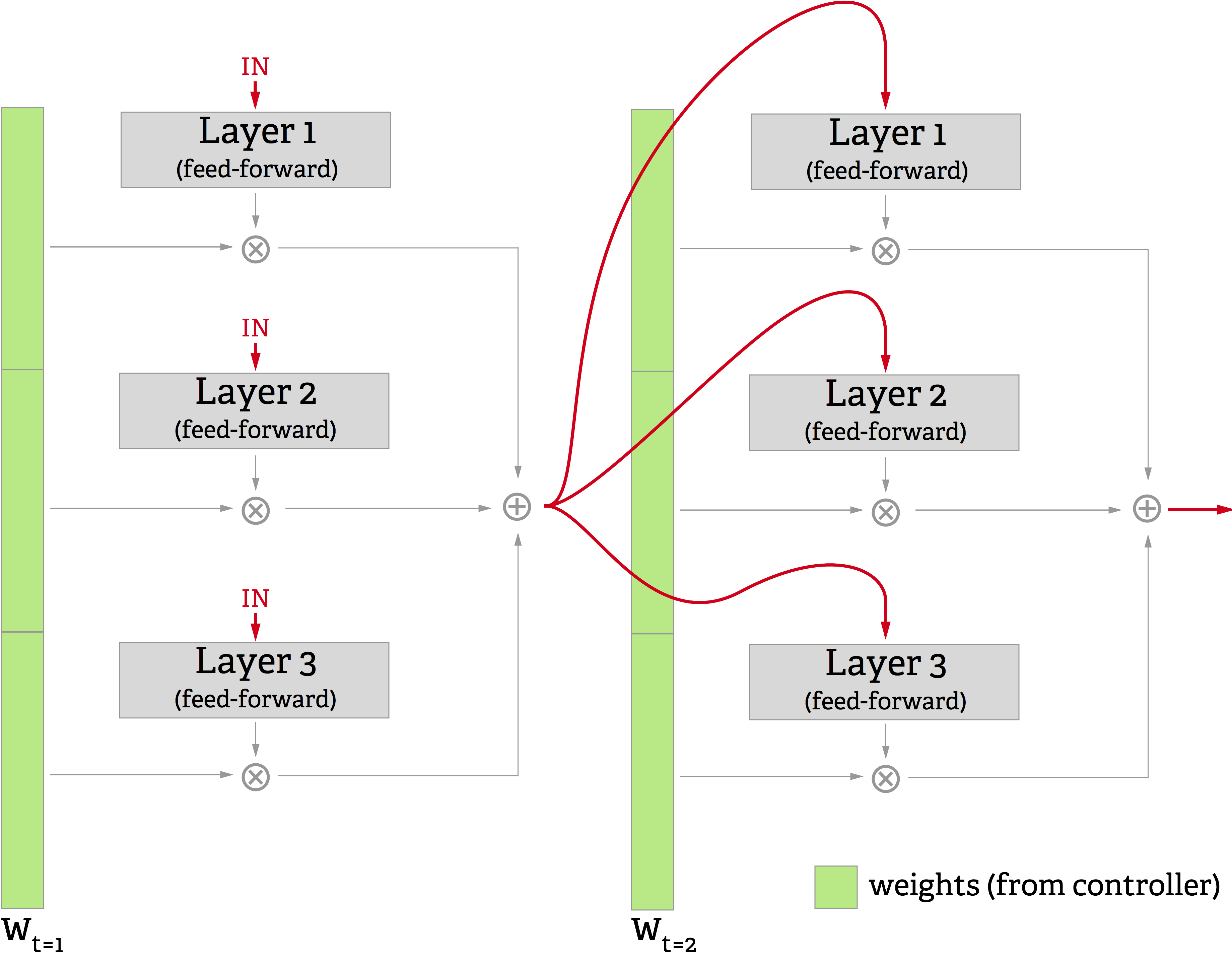}
\caption{\label{fig:multistep}\textbf{Multistep CFN.} A variant of the
design for using multiple timesteps (in this case, two) to calculate
each output.}
\end{figure}

One obvious question to consider about this model is, ``What happens if
the correct output function at a timestep is not computable with a
linear combination of single-layer networks?'' After all, there are
functions computable by a polynomial-width network of depth \texttt{k}
that require exponential width to compute with a network of depth
\texttt{k-1} (Hastad \protect\hyperlink{ref-hastad1986almost}{1986}).

To address this question, the system could be run for a predefined
number of steps between outputs. That is,

\begin{enumerate}
\def\labelenumi{\arabic{enumi}.}
\tightlist
\item
  Feed the system the input for ``real-world'' time \texttt{t} and save
  the output
\item
  Repeat \texttt{k} times:

  \begin{enumerate}
  \def\labelenumii{\alph{enumii}.}
  \tightlist
  \item
    Feed the system its own most recent output
  \end{enumerate}
\item
  Take the \texttt{kth} output of the system as its answer for time
  \texttt{t}
\item
  Repeat from 1. for time \texttt{t+1}
\end{enumerate}

This amounts to making the network deeper, in that more layers of
computation and nonlinearity lie between the input and the output. This
gives it the same computational power of a \texttt{k}-depth model.

While correctly learning the full latent factorization of data from
deeply-composed observations may be quite difficult, we can persist in
the theme of continuation learning and train such a system with a
curriculum, learning first to do one-step computations, then two, three,
and so forth until we have quite a complex model while still preserving
the disentangled nature of our representation. Various experiments done
by (Bengio et al. \protect\hyperlink{ref-bengio2009curriculum}{2009},
Zaremba and Sutskever
(\protect\hyperlink{ref-zaremba2014learning}{2014}), Reed and Freitas
(\protect\hyperlink{ref-reed2015neural}{2015})), among many others, have
shown using curricula with increasing complexity to be extremely
effective.

\subsubsection{Complex metadata}\label{complex-metadata}

While in the experiments described here the metadata given to the
controller is quite simple, the possibilities for the future are rich.
In the most straightforward case, the controller could read in source
code for a program one character at a time, then predict a sequence of
activations to compose the program it has read out of primitive
functions in the manner described in Sec.~\ref{sec:multistep}.

Even richer representations are also possible. The controller might for
example look at two images, then predict a series of structured
transformations to turn one into the other.

\subsubsection{Non-differentiable
functions}\label{non-differentiable-functions}

One unexplored capability of the controller-function network is that the
controller can produce weightings over ``functions'' or actions which
are not differentiable, while still being trained itself by
backpropagation. This is due to the structure of the output function:

\[Out = \sum_{i=1}^F C(x)_ {i} f_i(x))\]

\[\frac{\delta Out}{\delta C_i} = f_i(x)\]

The gradient of the controller does not depend on the gradient of the
functions. As a result, the functions that the controller uses could be
non-differentiable, ordinary handwritten programming functions. They
could even be \emph{actions} in the world, as long as those actions have
a continuous representation, e.g. \texttt{tighten\_left\_calf(force)}.

This convenient fact also has computational implications; if one were to
construct a hierarchy of CFNs, such that a higher-level CFN provides
weights on the functions computed by lower-level ones, all gradients
could be computed simultaneously and independently. This provides an
interesting avenue for future research in hierarchical computation.

\section{Unification}\label{unification}

Perhaps the most exciting direction for future work is the bringing
together of these two different techniques in order to build a
completely unsupervised system for inferring the factors of variation in
the world and using that structure to represent and predict videos.

Instead of using our knowledge of the active transformations over a
short video, we could train a controller network to \emph{weight} many
different transformations for each frame. Then, using the same
continuation methods described in this paper, we could sharpen those
weightings gradually over the course of training, resulting in a
representation of real video which is extremely sparse. Such a
representation might be able to capture the true generative structure of
simple scenes without any supervision at all, learning dimensions of
variation which are most common in the real world.

I look forward to continuing my work on building systems which can
understand the factors of variation in the world, and I hope that this
thesis provides some small help and inspiration to others in the field.

\chapter*{References}\label{references}
\addcontentsline{toc}{chapter}{References}

\hypertarget{refs}{}
\hypertarget{ref-aubry2014seeing}{}
Aubry, Mathieu, Daniel Maturana, Efros Alexei, Bryan Russell, and Josef
Sivic. 2014. ``Seeing 3D Chairs: Exemplar Part-Based 2D-3D Alignment
Using a Large Dataset of CAD Models.'' In \emph{CVPR}.

\hypertarget{ref-bengio2009learning}{}
Bengio, Yoshua. 2009. ``Learning Deep Architectures for AI.''
\emph{Foundations and Trends in Machine Learning} 2 (1). Now Publishers
Inc.: 1--127.

\hypertarget{ref-bengio2012deep}{}
---------. 2012. ``Deep Learning of Representations for Unsupervised and
Transfer Learning.'' \emph{Unsupervised and Transfer Learning Challenges
in Machine Learning} 7: 19.

\hypertarget{ref-bengio2013representation}{}
Bengio, Yoshua, Aaron Courville, and Pascal Vincent. 2013.
``Representation Learning: A Review and New Perspectives.''
\emph{Pattern Analysis and Machine Intelligence, IEEE Transactions on}
35 (8). IEEE: 1798--1828.

\hypertarget{ref-bengio2009curriculum}{}
Bengio, Yoshua, Jérôme Louradour, Ronan Collobert, and Jason Weston.
2009. ``Curriculum Learning.'' In \emph{Proceedings of the 26th Annual
International Conference on Machine Learning}, 41--48. ACM.

\hypertarget{ref-burda2015importance}{}
Burda, Yuri, Roger Grosse, and Ruslan Salakhutdinov. 2015. ``Importance
Weighted Autoencoders.'' \emph{ArXiv Preprint ArXiv:1509.00519}.

\hypertarget{ref-cohen2014learning}{}
Cohen, Taco, and Max Welling. 2014. ``Learning the Irreducible
Representations of Commutative Lie Groups.'' \emph{ArXiv Preprint
ArXiv:1402.4437}.

\hypertarget{ref-desjardins2012disentangling}{}
Desjardins, Guillaume, Aaron Courville, and Yoshua Bengio. 2012.
``Disentangling Factors of Variation via Generative Entangling.''
\emph{ArXiv Preprint ArXiv:1210.5474}.

\hypertarget{ref-dosovitskiy2015learning}{}
Dosovitskiy, A., J. Springenberg, and T. Brox. 2015. ``Learning to
Generate Chairs with Convolutional Neural Networks.''
\emph{ArXiv:1411.5928}.

\hypertarget{ref-erhan2009visualizing}{}
Erhan, Dumitru, Yoshua Bengio, Aaron Courville, and Pascal Vincent.
2009. ``Visualizing Higher-Layer Features of a Deep Network.''
\emph{Dept. IRO, Université de Montréal, Tech. Rep} 4323.

\hypertarget{ref-glorot2011domain}{}
Glorot, Xavier, Antoine Bordes, and Yoshua Bengio. 2011. ``Domain
Adaptation for Large-Scale Sentiment Classification: A Deep Learning
Approach.'' In \emph{Proceedings of the 28th International Conference on
Machine Learning (ICML-11)}, 513--20.

\hypertarget{ref-goodfellow2013empirical}{}
Goodfellow, Ian J, Mehdi Mirza, Da Xiao, Aaron Courville, and Yoshua
Bengio. 2013. ``An Empirical Investigation of Catastrophic Forgeting in
Gradient-Based Neural Networks.'' \emph{ArXiv Preprint ArXiv:1312.6211}.

\hypertarget{ref-graves2014neural}{}
Graves, Alex, Greg Wayne, and Ivo Danihelka. 2014. ``Neural Turing
Machines.'' \emph{ArXiv Preprint ArXiv:1410.5401}.

\hypertarget{ref-hastad1986almost}{}
Hastad, Johan. 1986. ``Almost Optimal Lower Bounds for Small Depth
Circuits.'' In \emph{Proceedings of the Eighteenth Annual ACM Symposium
on Theory of Computing}, 6--20. ACM.

\hypertarget{ref-he2015delving}{}
He, Kaiming, Xiangyu Zhang, Shaoqing Ren, and Jian Sun. 2015. ``Delving
Deep into Rectifiers: Surpassing Human-Level Performance on Imagenet
Classification.'' \emph{ArXiv Preprint ArXiv:1502.01852}.

\hypertarget{ref-hinton2011transforming}{}
Hinton, Geoffrey E, Alex Krizhevsky, and Sida D Wang. 2011.
``Transforming Auto-Encoders.'' In \emph{Artificial Neural Networks and
Machine Learning--ICANN 2011}, 44--51. Springer.

\hypertarget{ref-hinton2006fast}{}
Hinton, Geoffrey, Simon Osindero, and Yee-Whye Teh. 2006. ``A Fast
Learning Algorithm for Deep Belief Nets.'' \emph{Neural Computation} 18
(7). MIT Press: 1527--54.

\hypertarget{ref-hochreiter1997long}{}
Hochreiter, Sepp, and Jürgen Schmidhuber. 1997. ``Long Short-Term
Memory.'' \emph{Neural Computation} 9 (8). MIT Press: 1735--80.

\hypertarget{ref-jacobs1991task}{}
Jacobs, Robert A, Michael I Jordan, and Andrew G Barto. 1991. ``Task
Decomposition Through Competition in a Modular Connectionist
Architecture: The What and Where Vision Tasks.'' \emph{Cognitive
Science} 15 (2). Wiley Online Library: 219--50.

\hypertarget{ref-jacobs1991adaptive}{}
Jacobs, Robert A, Michael I Jordan, Steven J Nowlan, and Geoffrey E
Hinton. 1991. ``Adaptive Mixtures of Local Experts.'' \emph{Neural
Computation} 3 (1). MIT Press: 79--87.

\hypertarget{ref-jaderberg2015spatial}{}
Jaderberg, Max, Karen Simonyan, Andrew Zisserman, and others. 2015.
``Spatial Transformer Networks.'' In \emph{Advances in Neural
Information Processing Systems}, 2008--16.

\hypertarget{ref-kingma2013auto}{}
Kingma, Diederik P, and Max Welling. 2013. ``Auto-Encoding Variational
Bayes.'' \emph{ArXiv Preprint ArXiv:1312.6114}.

\hypertarget{ref-krizhevsky2012imagenet}{}
Krizhevsky, Alex, Ilya Sutskever, and Geoffrey E Hinton. 2012.
``Imagenet Classification with Deep Convolutional Neural Networks.'' In
\emph{Advances in Neural Information Processing Systems}, 1097--1105.

\hypertarget{ref-kulkarni2015picture}{}
Kulkarni, Tejas D, Pushmeet Kohli, Joshua B Tenenbaum, and Vikash
Mansinghka. 2015. ``Picture: A Probabilistic Programming Language for
Scene Perception.'' In \emph{Proceedings of the IEEE Conference on
Computer Vision and Pattern Recognition}, 4390--9.

\hypertarget{ref-kulkarni2014inverse}{}
Kulkarni, Tejas D, Vikash K Mansinghka, Pushmeet Kohli, and Joshua B
Tenenbaum. 2014. ``Inverse Graphics with Probabilistic CAD Models.''
\emph{ArXiv Preprint ArXiv:1407.1339}.

\hypertarget{ref-kulkarni2014variational}{}
Kulkarni, Tejas D, Ardavan Saeedi, and Samuel Gershman. 2014.
``Variational Particle Approximations.'' \emph{ArXiv Preprint
ArXiv:1402.5715}.

\hypertarget{ref-kulkarni2015deep}{}
Kulkarni, Tejas D, Will Whitney, Pushmeet Kohli, and Joshua B Tenenbaum.
2015. ``Deep Convolutional Inverse Graphics Network.'' \emph{ArXiv
Preprint ArXiv:1503.03167}.

\hypertarget{ref-lecun1995convolutional}{}
LeCun, Yann, and Yoshua Bengio. 1995. ``Convolutional Networks for
Images, Speech, and Time Series.'' \emph{The Handbook of Brain Theory
and Neural Networks} 3361. Cambridge, MA: MIT Press.

\hypertarget{ref-lee2009convolutional}{}
Lee, Honglak, Roger Grosse, Rajesh Ranganath, and Andrew Y Ng. 2009.
``Convolutional Deep Belief Networks for Scalable Unsupervised Learning
of Hierarchical Representations.'' In \emph{Proceedings of the 26th
Annual International Conference on Machine Learning}, 609--16. ACM.

\hypertarget{ref-liang2010learning}{}
Liang, Percy, Michael I Jordan, and Dan Klein. 2010. ``Learning
Programs: A Hierarchical Bayesian Approach.'' In \emph{Proceedings of
the 27th International Conference on Machine Learning (ICML-10)},
639--46.

\hypertarget{ref-mahendran2014understanding}{}
Mahendran, Aravindh, and Andrea Vedaldi. 2014. ``Understanding Deep
Image Representations by Inverting Them.'' \emph{ArXiv Preprint
ArXiv:1412.0035}.

\hypertarget{ref-mansimov2015generating}{}
Mansimov, Elman, Emilio Parisotto, Jimmy Lei Ba, and Ruslan
Salakhutdinov. 2015. ``Generating Images from Captions with Attention.''
\emph{ArXiv Preprint ArXiv:1511.02793}.

\hypertarget{ref-mansinghka2013approximate}{}
Mansinghka, Vikash, Tejas D Kulkarni, Yura N Perov, and Josh Tenenbaum.
2013. ``Approximate Bayesian Image Interpretation Using Generative
Probabilistic Graphics Programs.'' In \emph{Advances in Neural
Information Processing Systems}, 1520--8.

\hypertarget{ref-mnih2015human}{}
Mnih, Volodymyr, Koray Kavukcuoglu, David Silver, Andrei A Rusu, Joel
Veness, Marc G Bellemare, Alex Graves, et al. 2015. ``Human-Level
Control Through Deep Reinforcement Learning.'' \emph{Nature} 518 (7540).
Nature Publishing Group: 529--33.

\hypertarget{ref-mobahi2015theoretical}{}
Mobahi, Hossein, and John W Fisher III. 2015. ``A Theoretical Analysis
of Optimization by Gaussian Continuation.'' In \emph{Twenty-Ninth AAAI
Conference on Artificial Intelligence}.

\hypertarget{ref-mottaghi2014role}{}
Mottaghi, Roozbeh, Xianjie Chen, Xiaobai Liu, Nam-Gyu Cho, Seong-Whan
Lee, Sanja Fidler, Raquel Urtasun, and Alan Yuille. 2014. ``The Role of
Context for Object Detection and Semantic Segmentation in the Wild.'' In
\emph{IEEE Conference on Computer Vision and Pattern Recognition
(CVPR)}.

\hypertarget{ref-nair2008analysis}{}
Nair, Vinod, Josh Susskind, and Geoffrey E Hinton. 2008.
``Analysis-by-Synthesis by Learning to Invert Generative Black Boxes.''
In \emph{Artificial Neural Networks-ICANN 2008}, 971--81. Springer.

\hypertarget{ref-neelakantan2015neural}{}
Neelakantan, Arvind, Quoc V Le, and Ilya Sutskever. 2015. ``Neural
Programmer: Inducing Latent Programs with Gradient Descent.''
\emph{ArXiv Preprint ArXiv:1511.04834}.

\hypertarget{ref-paysan2009face}{}
Paysan, P., R. Knothe, B. Amberg, S. Romdhani, and T. Vetter. 2009. ``A
3D Face Model for Pose and Illumination Invariant Face Recognition.''
\emph{Proceedings of the 6th IEEE International Conference on Advanced
Video and Signal Based Surveillance (AVSS) for Security, Safety and
Monitoring in Smart Environments}. Genova, Italy: IEEE.

\hypertarget{ref-radford2015unsupervised}{}
Radford, Alec, Luke Metz, and Soumith Chintala. 2015. ``Unsupervised
Representation Learning with Deep Convolutional Generative Adversarial
Networks.'' \emph{ArXiv Preprint ArXiv:1511.06434}.

\hypertarget{ref-ranzato2007unsupervised}{}
Ranzato, M, Fu Jie Huang, Y-L Boureau, and Yann LeCun. 2007.
``Unsupervised Learning of Invariant Feature Hierarchies with
Applications to Object Recognition.'' In \emph{Computer Vision and
Pattern Recognition, 2007. CVPR'07. IEEE Conference on}, 1--8. IEEE.

\hypertarget{ref-reed2015neural}{}
Reed, Scott, and Nando de Freitas. 2015. ``Neural
Programmer-Interpreters.'' \emph{ArXiv Preprint ArXiv:1511.06279}.

\hypertarget{ref-salakhutdinov2009deep}{}
Salakhutdinov, Ruslan, and Geoffrey E Hinton. 2009. ``Deep Boltzmann
Machines.'' In \emph{International Conference on Artificial Intelligence
and Statistics}, 448--55.

\hypertarget{ref-sukhbaatar2015end}{}
Sukhbaatar, Sainbayar, Jason Weston, Rob Fergus, and others. 2015.
``End-to-End Memory Networks.'' In \emph{Advances in Neural Information
Processing Systems}, 2431--9.

\hypertarget{ref-tang2012deep}{}
Tang, Yichuan, Ruslan Salakhutdinov, and Geoffrey Hinton. 2012. ``Deep
Lambertian Networks.'' \emph{ArXiv Preprint ArXiv:1206.6445}.

\hypertarget{ref-tenenbaum2000separating}{}
Tenenbaum, Joshua B, and William T Freeman. 2000. ``Separating Style and
Content with Bilinear Models.'' \emph{Neural Computation} 12 (6). MIT
Press: 1247--83.

\hypertarget{ref-theis2015generative}{}
Theis, Lucas, and Matthias Bethge. 2015. ``Generative Image Modeling
Using Spatial LSTMs.'' In \emph{Advances in Neural Information
Processing Systems}, 1918--26.

\hypertarget{ref-rmsprop}{}
Tieleman, T., and G. Hinton. 2012. ``Lecture 6.5 - Rmsprop, COURSERA:
Neural Networks for Machine Learning.''

\hypertarget{ref-tieleman2014optimizing}{}
Tieleman, Tijmen. 2014. ``Optimizing Neural Networks That Generate
Images.'' PhD thesis, University of Toronto.

\hypertarget{ref-vincent2010stacked}{}
Vincent, Pascal, Hugo Larochelle, Isabelle Lajoie, Yoshua Bengio, and
Pierre-Antoine Manzagol. 2010. ``Stacked Denoising Autoencoders:
Learning Useful Representations in a Deep Network with a Local Denoising
Criterion.'' \emph{The Journal of Machine Learning Research} 11. JMLR.
org: 3371--3408.

\hypertarget{ref-williams1992simple}{}
Williams, Ronald J. 1992. ``Simple Statistical Gradient-Following
Algorithms for Connectionist Reinforcement Learning.'' \emph{Machine
Learning} 8 (3-4). Springer: 229--56.

\hypertarget{ref-yang2015weakly}{}
Yang, Jimei, Scott E Reed, Ming-Hsuan Yang, and Honglak Lee. 2015.
``Weakly-Supervised Disentangling with Recurrent Transformations for 3d
View Synthesis.'' In \emph{Advances in Neural Information Processing
Systems}, 1099--1107.

\hypertarget{ref-yosinski2014transferable}{}
Yosinski, Jason, Jeff Clune, Yoshua Bengio, and Hod Lipson. 2014. ``How
Transferable Are Features in Deep Neural Networks?'' In \emph{Advances
in Neural Information Processing Systems}, 3320--8.

\hypertarget{ref-zaremba2014learning}{}
Zaremba, Wojciech, and Ilya Sutskever. 2014. ``Learning to Execute.''
\emph{ArXiv Preprint ArXiv:1410.4615}.

\hypertarget{ref-zaremba2015reinforcement}{}
---------. 2015. ``Reinforcement Learning Neural Turing Machines.''
\emph{ArXiv Preprint ArXiv:1505.00521}.

\hypertarget{ref-zaremba2015learning}{}
Zaremba, Wojciech, Tomas Mikolov, Armand Joulin, and Rob Fergus. 2015.
``Learning Simple Algorithms from Examples.'' \emph{ArXiv Preprint
ArXiv:1511.07275}.

\hypertarget{ref-zeiler2014visualizing}{}
Zeiler, Matthew D, and Rob Fergus. 2014. ``Visualizing and Understanding
Convolutional Networks.'' In \emph{Computer Vision--ECCV 2014}, 818--33.
Springer.

\begin{singlespace}

\end{singlespace}

\end{document}